\documentclass{article}
\usepackage{iclr2022_workshop,times}

\usepackage{amsmath,amsfonts,bm}

\def\eqref#1{equation~\ref{#1}}

\def\1{\bm{1}}

\DeclareMathAlphabet{\mathsfit}{\encodingdefault}{\sfdefault}{m}{sl}
\SetMathAlphabet{\mathsfit}{bold}{\encodingdefault}{\sfdefault}{bx}{n}

\DeclareMathOperator{\Tr}{Tr}

\usepackage{graphicx}
\usepackage{wrapfig}
\usepackage{caption}
\usepackage{subcaption}
\usepackage[colorlinks=true, allcolors=blue]{hyperref}
\usepackage[utf8]{inputenc}
\usepackage{amsmath,amssymb}
\usepackage{multirow}
\usepackage{parskip}
\usepackage[nottoc]{tocbibind}
\usepackage{enumitem}
\usepackage{pifont}
\usepackage{url}
\usepackage{mathtools}

\title{An extensible Benchmarking Graph-Mesh dataset for studying Steady-State Incompressible Navier-Stokes Equations}

\author{Florent Bonnet\textsuperscript{1,2,3}, Jocelyn Ahmed Mazari\textsuperscript{3}, Thibaut Munzer\textsuperscript{3}, Pierre Yser\textsuperscript{3}, Patrick Gallinari\textsuperscript{1,4} \\
\textsuperscript{1} Sorbonne Université, CNRS, ISIR, F-75005 Paris, France\\
\textsuperscript{2} École Normale Supérieure Paris Saclay, France \\
\textsuperscript{3} Extrality, 75002 Paris, France\\
\textsuperscript{4} Criteo AI Lab, Paris, France\\
\texttt{\{florent, ahmed, thibaut,pierre\}@extrality.ai}\\
\texttt{patrick.gallinari@sorbonne-universite.fr}
}

\iclrfinalcopy 
\begin{document}

\maketitle

\begin{abstract}

Recent progress in \emph{Geometric Deep Learning} (GDL) has shown its potential to provide powerful data-driven models. This gives momentum to explore new methods for learning physical systems governed by \emph{Partial Differential Equations} (PDEs) from Graph-Mesh data. However, despite the efforts and recent achievements, several research directions remain unexplored and progress is still far from satisfying the physical requirements of real-world phenomena. One of the major impediments is the absence of benchmarking datasets and common physics evaluation protocols. In this paper, we propose a 2-D graph-mesh dataset to study the airflow over airfoils at high Reynolds regime (from $10^6$ and beyond). We also introduce metrics on the stress forces over the airfoil in order to evaluate GDL models on important physical quantities. Moreover, we provide extensive GDL baselines. 
\end{abstract}

\section{Introduction and motivation}\label{sec:intro}

The conception of cars, planes, rockets, wind turbines, boats, etc. requires the analysis of surrounding physical fields. Measuring them experimentally by building prototypes is time-consuming, computationally expensive, and sometimes dangerous. Having a measurement in a particular point or reproducing a complex configuration is sometimes impossible during the test phase. Virtual testing allows us to tackle many of these constraints and to test configurations that could not be possible in reality.
Hence, numerical simulations are crucial for modeling physical phenomenas and are especially used in \emph{Computational Fluid Dynamics} (CFD) to solve Navier-Stokes equations. 

At high Reynolds number, these equations involve a complex dissipation process that cascade from large length scales to small ones which make direct resolutions challenging. Traditional CFD frameworks rely on turbulence models and the power of intensive parallel computations to simulate and analyze fluid dynamics. Despite the efficiency of existing tools, fluid numerical simulations are still computationally expensive and can take several weeks to converge to an accurate solution.

Nevertheless, a huge quantity of data can be extracted from numerical simulation solutions and are today available to assess a new way to recover fluid flow solutions. These data could be exploited to explore the potential of data-driven models in approximating Navier-Stokes PDEs by capturing highly non-linear phenomena. The framework of \emph{Deep Learning} (DL) is one of the successful data-driven methods that have drawn lots of attention recently \citep{NIPS2012_4824,7780582,10.1145/3439726,pmlr-v48-amodei16}. DL methods are particularly interesting thanks to their universal approximation properties \citep{NEURIPS2020_2000f632,10.5555/70405.70408}; capable of approximating a wide range of continuous functions, which makes it a relevant candidate to tackle physics problems while leading to new perspectives for CFD. PDE and DL offer complementary strengths: the modeling power, interpretability, and the accuracy of differential equations solutions, as well as the approximation capabilities and inference speed of DL methods.\\
In practice, CFD solvers operate on meshes to solve Navier-Stokes equations. However, standard DL models such as \emph{Convolutional Neural Networks} (CNNs) \citep{NIPS2012_4824} achieve learning on regular grid data. Hence, CNNs are not designed to operate directly on meshes but several methods based on grid approaches \citep{um2020sol,thuerey2020deepFlowPred,mohan2020embedding,wandel2021learning,10.1145/3392717.3392772,gupta2021multiwaveletbased} have been proposed to approximate the functional space of PDEs. Unfortunately, they cannot correctly infer the physical fields close to obstacles, which make it difficult to correctly compute stress forces at the surface of a geometry. In addition to grid approaches, PDE-based supervised learning methods such as \emph{Physics-Informed Neural Networks} (PINNs) \citep{raissi2019physics} have emerged and could help to solve the physical fields in the entire continuous domain but they are difficult to train \citep{Wang2021UnderstandingAM} and are restricted to solving one predefined PDE. Recently, DL on unstructured data has been categorized under the name of \emph{Geometric Deep Learning} (GDL) \citep{DBLP:journals/spm/BronsteinBLSV17}. It consists in designing geometrical and compositional inductive biases in DL, reflecting the rich and complex structure in the data. \emph{Graph Neural Networks} (GNNs) \citep{1555942,4700287,Li2016GatedGS,Kipf:2016tc,10.5555/3305381.3305512,pmlr-v80-sanchez-gonzalez18a} are part of this category and several works on GNNs for approximating PDEs have been proposed \citep{pfaff2021learning,xu2021conditionally,brandstetter2022message}, including solver in the loop method \citep{belbute_peres_cfdgcn_2020} and graph neural operator methods   \citep{GKN,MGKN}. An important advantage of GDL techniques is their ability to predict quantities over arbitrary shapes without requiring their voxelizations. In our case, this allows accurate computations of stress forces over airfoils.

In this work, we propose a benchmarking graph-mesh dataset for studying the problem of 2-D steady-state incompressible \emph{Reynolds-Averaged Navier-Stokes} (RANS) equations along with an evaluation protocol, especially the computation of stress forces at the surface of the airfoils. Finally, we conduct extensive experiments to give insight about the potential of DL for solving physics problems. We would like to mention that we identified in the literature one similar work but based on structured grid data, that proposes a set of benchmarks to study physical systems \citep{otness2021an} with classical DL approaches.

\section{Dataset construction and description}
\begin{figure}
      \centering
    \begin{subfigure}{0.4\textwidth}
      \centering
      \includegraphics[width=\linewidth]{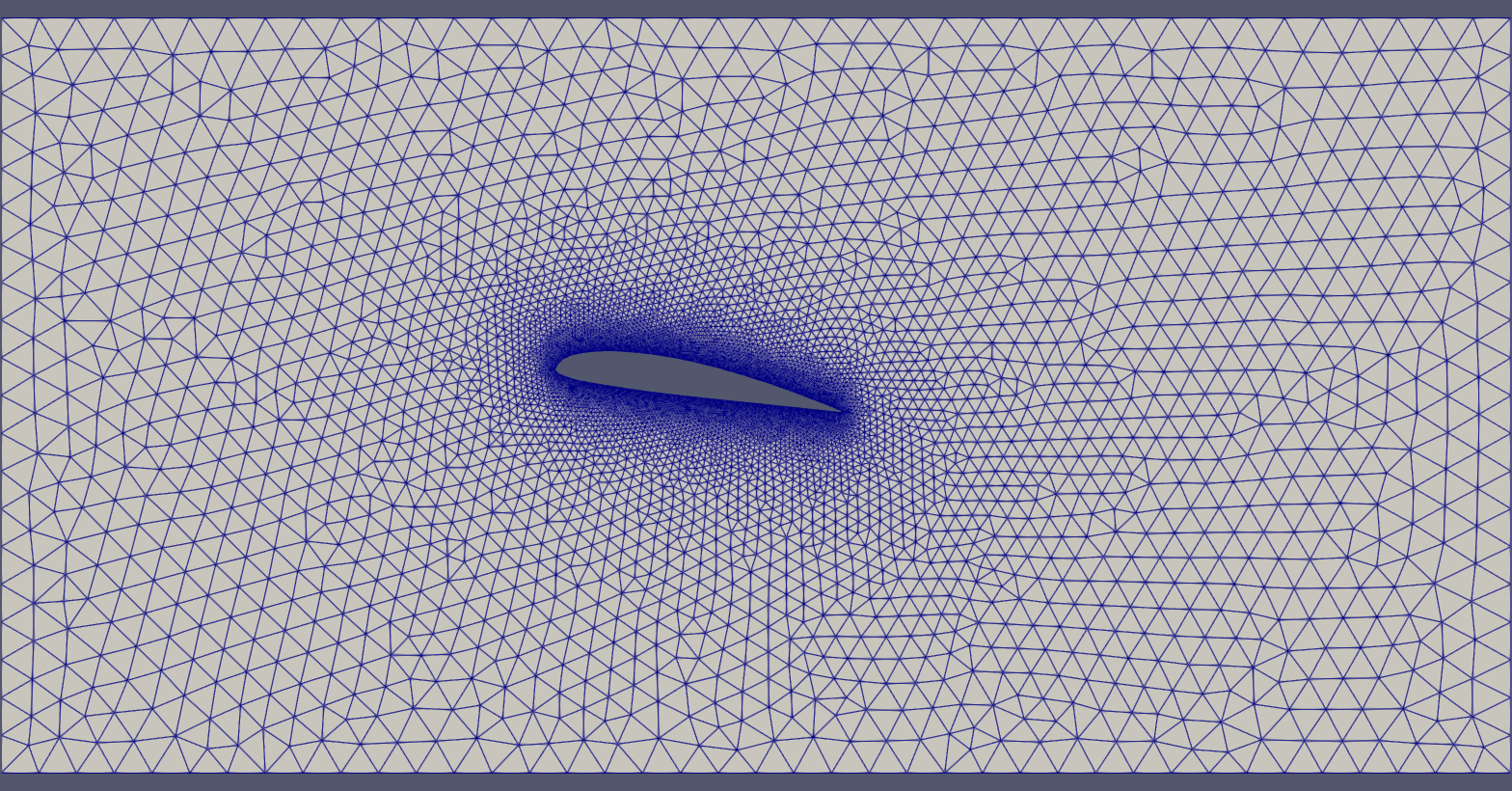} 
      \caption{}
    \end{subfigure}
    $\xRightarrow[\text{\bf GNN}]{\text{\bf CFD Solver}}$
    \begin{subfigure}{0.4\textwidth}
      \centering
      \includegraphics[width=\linewidth]{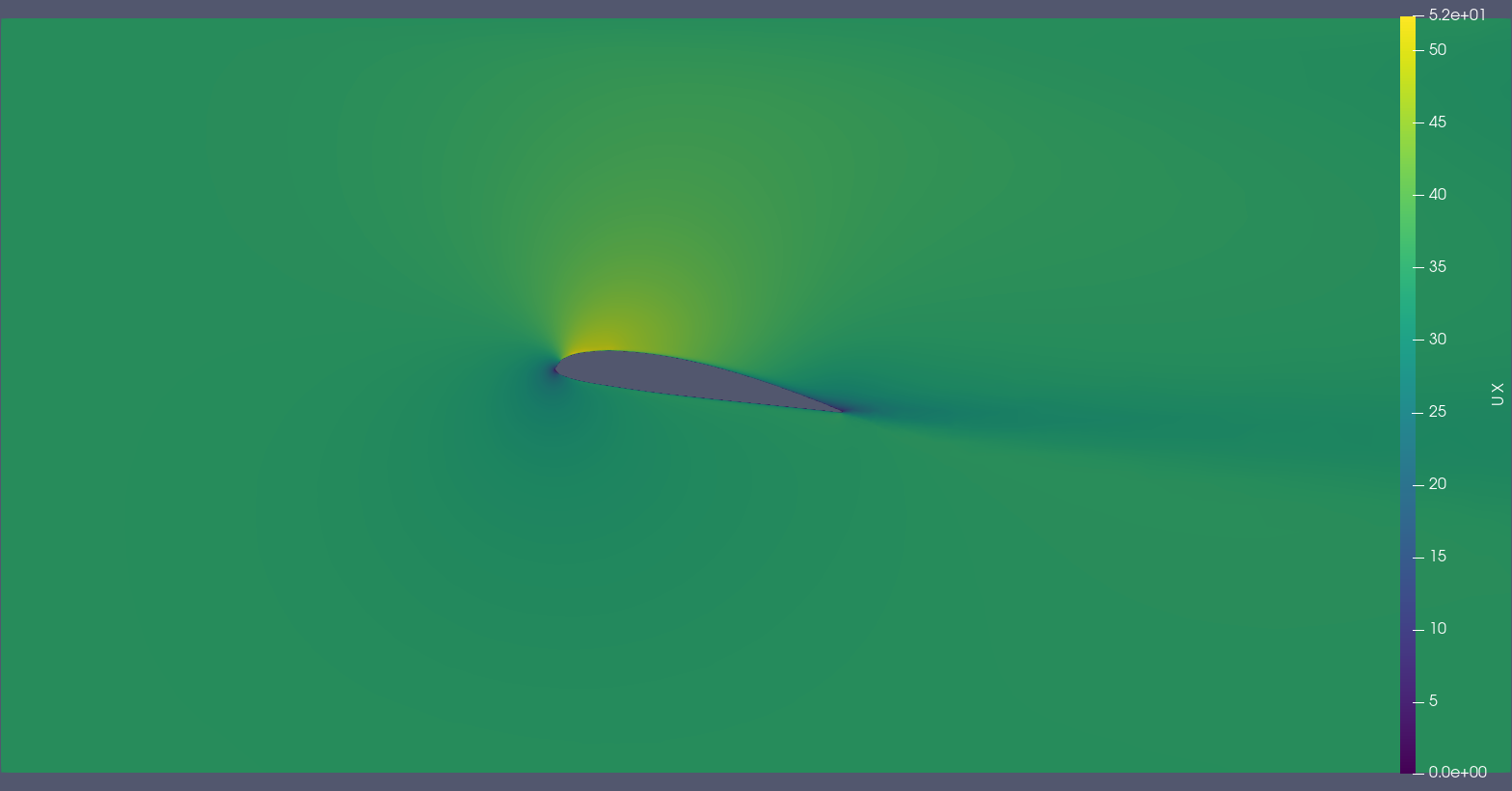}
      \caption{}
    \end{subfigure}

    \caption{(a) Airfoil mesh. (b) $x$-velocity field. The goal of a GNN is to produce outputs that are as accurate as those of a CFD solver while drastically reducing the computational time.
    }
    \label{fig:cfd_dl}
\end{figure}

We chose OpenFOAM \citep{Jasak07openfoam:a}, an open-source CFD software to run our simulations. We use the steady-state solver simpleFOAM to run Reynolds-Averaged-Simulation with the Spalart-Allmaras model \citep{SPALART92} for the turbulence modeling. These simulations were done for multiple airfoil geometries\footnote{Geometries from the National Advisory Committee for Aeronautics (NACA).}.
We distinguish three different sets; namely  training set, validation set, and test set that are taken from the same distribution. The overall 2-D dataset statistics are reported in Table \ref{tab:dataset_stat}. 
In this work, a geometry represents a shape of an airfoil and an angle of attack. For each geometry, we define a compact domain around it, on which a tetrahedral mesh is built (see Appendix \ref{ap:graph_mesh_gen}). Inlet velocity is uniformly sampled between 10 and 50 meters per second which corresponds to a Reynolds number between $10^{6}$ and $5\cdot10^6$ and a Mach number between 0.03 and 0.15. The characteristic length of the proposed airfoils is 1 meter, the kinematic viscosity of air is approximated by $10^{-5}$. The meshes and boundary conditions are fed to the CFD solver to run the simulations. Its outputs are four fields that represent the targets in our supervised learning task, namely the $x$ and $y$ components of the velocity vector, the pressure and the turbulent viscosity. All those information are wrapped up in a ready-to-use way via the framework of \emph{PyTorch Geometric} (PyG). Figure \ref{fig:cfd_dl} depicts an example of the input/outputs of CFD solvers and GNNs.

The related details to the construction of the dataset and the normalization procedure are described in the Appendices \ref{ap:graph_mesh_gen} and \ref{ap:stat_preprocessing}.

\begin{table}[t!]
    \caption{Properties of the different sets. An interval means that the parameter for the related quantity has been drawn from a uniform distribution over this interval.}
    \resizebox{0.99\textwidth}{!}{
 \centering
    \begin{tabular}{|c|c||c|c|c||c|c|c||c|c|c|c|}
     \bf Sets& \bf \#Samples &  \multicolumn{6}{|c||}{\bf Graph-mesh parameters} &     \multicolumn{4}{|c}{\bf Physical parameters}
        \\
         \multirow{2}{*}{}& \multirow{2}{*}{} &  \multicolumn{3}{|c||}{\bf \#nodes/sample} &     \multicolumn{3}{|c||}{\bf \#edges/sample}& \bf Angle of attack & \bf Inlet velocity& \bf Reynolds & \bf Mach \\
                 
         &  & \bf Mean  &\bf  Min &\bf  Max &\bf  Mean &\bf Min & \bf Max & ($deg$) & ($m\cdot s^{-1}$) &\bf number & \bf number\\
           &  &   & & & & &  & & & &\\
            \hline
                  &  &   & & & & & & & & &\\
            Train & 180 & 13012 & 9585 & 17102 & 75803 & 55804 & 99818 &  $[-0.3,0.3]$ & $[10,50]$ & $[10^6, 5\cdot10^6]$ & $[0.03,0.15]$\\
                  &&  &   & & & & &  & & &\\
          \hline
                 &  &  & & & & & &  & & &\\
            Validation & 20 & 12694 &10028 &15587  & 73948 & 58380 & 90934 &  $[-0.3,0.3]$ & $[10,50]$ & $[10^6, 5\cdot10^6]$ & $[0.03,0.15]$\\
                  &  & &  & & & & &  & & &\\
                    \hline
                          &  & &  & & & & &  & & &\\
            Test & 30 & 13519 & 10225 & 16193 & 78746 & 59570 & 94368 & $[-0.3,0.3]$ & $[10,50]$ & $[10^6, 5\cdot10^6]$ & $[0.03,0.15]$ \\
  
    \end{tabular}
    }
    
    \label{tab:dataset_stat}
\end{table}

\section{Task definition and physical metrics}\label{sec:physics_metrics_task}

The incompressible steady-state RANS equations used for this dataset

can be expressed as:
\begin{align}\label{eq:rans_eq}
    \begin{cases}
        (U\cdot \nabla)U = -\frac{1}{\rho}\nabla p + (\nu + \nu_t) \Delta U \\
        \nabla\cdot U = 0
    \end{cases}
\end{align}
where $U$ is the time-average velocity, $\rho$ the density of the fluid, $p$ an effective time-average pressure, $\nu$ the kinematic viscosity and $\nu_t$ the kinematic turbulent viscosity. Boundary conditions are added along with the Spalart-Allmaras turbulence equation to close the problem. 

The loss $\mathcal{L}$ used in this work is the sum of two terms, a loss over the volume $\mathcal{L}_{\mathcal{V}}$ and a loss over the surface $\mathcal{L}_{\mathcal{S}}$:
\begin{equation}\label{eq:global_loss}
\centering
    \mathcal{L}:={\underbrace{\frac{1}{|\mathcal{V}|}\sum_{i\in \mathcal{V}} \|f_\theta(x_i) - y_i\|_2^2}_{\text{$\mathcal{L}_\mathcal{V}$ loss over the volume}}}+  \lambda{\underbrace{\frac{1}{|\mathcal{S}|}\sum_{i\in \mathcal{S}} \|f_\theta(x_i) - y_i\|_2^2}_{\text{$\mathcal{L}_\mathcal{S}$ loss over the surface}}}
\end{equation}
where $\mathcal{V}$, $\mathcal{S}$ are respectively the set of the indices of the nodes that lie in the volume and on the surface respectively, $x_i \in \mathbb{R}^{4}$ is the input at node $i$ containing the 2-D spatial coordinates, the velocity of the flow at the inlet and the Euclidean distance function between the node and the airfoil, $y_i \in \mathbb{R}^{4}$ the targets at node $i$ containing the 2-D velocity, the pressure and the kinematic turbulent viscosity at this node and $f_\theta$ the model used.
The coefficient $\lambda$ is used to balance the weight of the error at the surface of the geometry and over the volume\footnote{In this work $\lambda$ is set to 1.}. We have to emphasize that this loss is not necessarily a good proxy when it comes, for instance, to compute the wall shear stress or to ensure that the inferred velocity field is divergence free.

At the end of the training, we compute the global \emph{Wall Shear Stress} (WSS) and \emph{Wall Pressure} (WP), see appendix \ref{sec:drag_lift} for the definition of those quantities. The WSS and WP allow to recover the stress forces at the surface of the geometry and the integral geometry forces

which are the \emph{drag} and the \emph{lift}. To the best of our knowledge, this is the first work that proposes an evaluation protocol to assess DL models not only on the quantity regressed but also on more meaningful
metrics

for real-world problems that require geometries.

In the literature most of the models are experimented over the volume, except the recent work \citep{Suk/et/al2022} that proposes a model to regress WSS only on a surface mesh. It is not the direction we chose for our baselines, as we want our model to output only the velocity, pressure and turbulent viscosity fields in order to stick with the form of the RANS equations.

From a given geometry and physical parameters, the ultimate goal is to accurately regress the velocity, pressure, and turbulent viscosity fields, as well as to compute the stress forces acting on this geometry. Because data comes in the form of graphs, the task is a real challenge as most of the works focus on regular girds, though a few interesting solutions on meshes are currently emerging as mentioned in section \ref{sec:intro}. Furthermore, our inlet velocity corresponds to high Reynolds, from $10^{6}$ to $5\cdot10^6$, which is closer to real-world problems while precedent works focus on Reynolds of some orders of magnitude smaller.

\section{Experiments}
\textbf{Task definition.} For our baselines, we chose to regress the unknown fields involved in the RANS equations \ref{eq:rans_eq} and to compute the stress forces as a post-processing step.

\textbf{Controlling the numerical complexity.} The number of nodes and edges in CFD meshes are very high (especially in 3-D cases). Hence, the methods used have to be robust to downsampling to better generalize to more complex problems. This implies that we cannot directly use a CFD mesh as input. One way to overcome this problem is to uniformly draw a subsampling of points in the entire mesh and to build a graph on these point clouds. During the training, at each sample and at each epoch, 1600 different nodes are sampled which represent 10\% to 20\% of the total number of nodes (depending on the simulation). If needed, a graph is built over this point cloud by connecting the nodes that are closer than a fixed Euclidean distance. We call this graph, \emph{radius graph}. We chose the radius of our graphs to be 0.1 and we set the maximum number of neighbor points to 64 in order to limit the memory footprint. The entire normalized domain is roughly square of length 8 (in normalized unit). Hence, we connect only very locally and the resulting graphs are not necessarily connected. For the inference, we keep all the points of the CFD meshes, rebuild graphs of radius 0.1 and set the maximum number of neighbor points to 512.

\textbf{Baselines.} We apply several \emph{Geometric Deep Learning} (GDL) models to the dataset developed in this work. We distinguish two families of methods, \emph{single-scale} models and \emph{multi-scale} models. The former include GraphSAGE \citep{SAGE}, GAT \citep{GAT}, PointNet \citep{pointnet} and GNO \citep{GKN} while the latter is composed of Graph-Unet \citep{GraphUNet}, PointNet++ \citep{pointnet++}, and MGNO \citep{MGKN}. The GraphSAGE, GAT, GNO, Graph-Unet and MGNO are graph-based models while PointNet and PointNet++ act on point clouds. The GNO and MGNO belong to the class of neural operator methods that consist in learning a mapping between two Hilbert spaces. We use Adam \citep{kingma2014method} along with one-cycle cosine learning rate scheduler \citep{n.2018superconvergence} of maximum $3\cdot 10^{-3}$ to optimize our neural networks.

Details of the architectures are provided in the supplementary material along with an ablation study. We present in Table \ref{total_score} scores for the trained models evidencing the difficulty of the various baselines to perform well on all real-world metrics (see Appendix \ref{sec:result_discussion} for results discussion).

The MSE for the stress forces is computed with the unnormalized inferred quantities whereas the MSE over the surface and the volume is computed with normalized quantities (see  Appendix \ref{ap:stat_preprocessing}).
\begin{table}
    \caption{Scores in MSE of the different models over the test set, $\mathcal{L}_\mathcal{V}$ and $\mathcal{L}_\mathcal{S}$ are given in terms of normalized quantities and the integral geometry forces are given in terms of unnormalized quantities. Each model is trained 10 times. GNO* and MGNO* stand respectively for our modified GNO \citep{GKN} and MGNO \citep{MGKN} in comparison with GrapheSAGE \citep{SAGE}, GAT \citep{GAT}, PointNet \citep{pointnet}, Graph-Unet \citep{GraphUNet}, and PointNet++ \citep{pointnet++}. See Appendix \ref{sec:details_exp}  for the details of each model.}
    \resizebox{0.99\textwidth}{!}{
    \centering
    \begin{tabular}{|c|c|c|c|c|c|c|c|c|c|c|c|c}

        \bf Models & & \multicolumn{2}{c|}{\bf Local metrics} &\multicolumn{4}{c|}{\bf Integral geometry forces} & \bf \#Params & \bf Training & \bf Inference    \\
                          & &  &  &  &  &  &  & & \bf time & \bf time  \\
        & & $\mathcal{L}_\mathcal{V}$ & $\mathcal{L}_\mathcal{S}$ & x-WSS &   y-WSS & x-WP &  y-WP & & &\\
                & & $\left(\times 10^{-2}\right)$ & $\left(\times 10^{-2}\right)$ & $\left(\times 10^{-3}\right)$ & $\left(\times 10^{-4}\right)$ & $\left(\times 10\right)$ & $\left(\times 10^{2}\right)$ &  & (hh:mm.ss) & (sec)  \\
         \hline
                         & &  &  &  &  &  &  &  &  &\\
        \multirow{4}{*}{\bf Single-scale} & GraphSAGE & \textbf{2.97 $\pm$ 0.15} & \textbf{4.81 $\pm$ 0.27} & 5.19 $\pm$ 1.04 & 3.39 $\pm$ 0.66 & 4.42 $\pm$ 1.13 & 7.18 $\pm$ 1.81 & 29140 & 0:10.47 & 0.40 \\
         & GAT & 3.20 $\pm$ 0.29 & 35.9 $\pm$ 15.3 & 235 $\pm$ 172 & 16.5 $\pm$ 11.7 & \textbf{3.64 $\pm$ 1.01} & \textbf{6.59 $\pm$ 1.22} & 47924 &  0:14.56  & 0.84 \\
         & PointNet & 4.31 $\pm$ 0.25 & 8.33 $\pm$ 1.01 & 12.6 $\pm$ 3.75 & 4.82 $\pm$ 2.14 & 7.33 $\pm$ 1.31 & 20.7 $\pm$ 5.97 & 75180 & 0:09.56 & 0.40\\
        & GNO* & 3.02 $\pm$ 0.24 & 5.29 $\pm$ 0.27 & \textbf{5.15 $\pm$ 1.26} & \textbf{3.37 $\pm$ 1.17} & 5.91 $\pm$ 2.33 & 9.02 $\pm$ 1.97 & 23260 & 0:20.28 & 0.41 \\
                        & &  &  &  &  &  &  &  &  & \\
         \hline
                         & &  &  &  &  &  &  &  &  & \\
         \multirow{4}{*}{\bf Multi-scale} & Graph-Unet & 3.03 $\pm$ 0.67 & 4.98 $\pm$ 0.37 & 6.26 $\pm$ 1.66 & 3.71 $\pm$ 0.98 & 6.26 $\pm$ 1.66 & \textbf{7.35 $\pm$ 1.49} & 65756 & 0:21.03 & 0.42\\
         & PointNet++ & 3.61 $\pm$ 0.60 & 6.39 $\pm$ 0.44 & 5.99 $\pm$ 2.67 & 4.43 $\pm$ 1.36 & 6.75 $\pm$ 2.47 & 11.7 $\pm$ 3.94 & 4046156 & 0:54.20 & 0.77 \\
        & MGNO* & \textbf{1.83 $\pm$ 0.12} & \textbf{4.03 $\pm$ 0.28} & \textbf{3.94 $\pm$ 1.39} & \textbf{1.80 $\pm$ 0.34} & \textbf{2.99 $\pm$ 0.69} & 8.24 $\pm$ 1.36 & 75484 & 5:14.38 & 2.01 \\
        
    \end{tabular}
    }
    \label{total_score}
\end{table}

\section{Conclusion}
In this work, we developed a preliminary version of a graph-mesh dataset to study 2-D steady-state incompressible RANS equations along with physics-based evaluation protocol. We also provided a set of appropriate baselines to illustrate the potential of GDL to partially replace CFD solvers leading to new perspectives for design and shape optimization processes. We proposed a new way of measuring the performance of DL models and we underlined the importance to use unstructured data in order to have access to more physical quantities such as stress forces. We argue that there is no step forward in a Machine Learning task without a decent set of data and well predefined metrics. This work is a first step to propose elements for the task of quantitative and real-world physics problems.

\textbf{Future works}
will be dedicated to a brand new version of this dataset with more precision in the CFD meshing process and proof of convergence of the CFD simulations. Moreover, we would like to increase the level of difficulty of the test set by proposing test data out of the training distribution (for Reynolds slightly higher or lower from the training data). We also consider the importance of developing similar data for compressible, unsteady flows, and 3-D cases.

\section{Broader impact}
This work could be used to:
\begin{enumerate}
    \item experiment new GDL models in this area,
    \item study the capabilities of DL to capture physical phenomena relying on our physics-based evaluation protocol,
    \item give insights to establish new research directions for numerical simulation and ML following the behaviors that will be observed in our quantitative and qualitative results (as well as in the next version of the dataset) with the physical metrics,
    \item extend graph benchmarking datasets and applications of GNNs to physics problems,
    \item build surrogate solvers to help CFD engineers to optimize design cycles and iterate efficiently as much as needed on their designs,
\end{enumerate}

\section{Reproductibility Statement}\label{sec:repro}
We provide a  \href{https://github.com/Extrality/ICLR_NACA_Dataset_V0}{GitHub repository} to reproduce the experiments and a \href{https://data.isir.upmc.fr/extrality/2D_RANS_NACA_Dataset.zip}{link} to download the dataset. The experiments have been done with a NVIDIA GeForce RTX 3090 24Go.

\newpage
\bibliography{iclr2022_workshop}
\bibliographystyle{iclr2022_workshop}
\newpage
\appendix

\section{Description of software}\label{ap:tools_def}
In this section, we describe the tools that we have used in this work to build the dataset, make the visualizations, and train the models. This work makes use of computational fluid dynamics (CFD) and ML tools.

\textbf{OpenFOAM} \citep{Jasak07openfoam:a} stands for \emph{Open-source Field Operation And Manipulation}, a C++ software for developing custom numerical solvers to study continuum mechanics and CFD problems. In this work, we have used version 8.0 of OpenFOAM to make our simulations. OpenFOAM is released as free and open-source software under the \emph{GNU General Public Licence}.

\textbf{Gmsh} \citep{Geuzaine2009} is an open-source meshing tool based on 3-D finite element mesh generator with a built-in CAD engine. It supports an Application Programming Interface (API) in four languages: C , C++, Python and Julia. This tool is also able to build meshes in 2-D. We have used version 4.9.3 of Gmsh in this work. Gmsh is released as free and open-source software under the \emph{GNU General Public Licence}.

\textbf{ParaView} \citep{10.5555/2789330} is an open-source visualization tool designed to explore and visualize efficiently large data using quantitative and qualitative metrics. ParaView runs on distributed and shared memory parallel and single processor systems. In this work, we have used it to visualize the following: point clouds, meshes, the predicted (as well as the ground truth) physical fields. We have used version 5.7.0 of ParaView in this work. ParaView  is released  as free and open-source software under the \emph{Berkeley Software Distribution License}.

\textbf{PyVista} \citep{sullivan2019pyvista} is an open-source tool based on a handy interface for the Visualization ToolKit (VTK). It is simple to use in interaction with NumPy \citep{citeulike:9919912} and other Python libraries. It is mainly used for mesh analysis. In this work, we use PyVista to build the inputs of our DL models. We have used version 0.33.0 of PyVista in this work. PyVista is released as free and open-source software under the \emph{MIT License}.

\textbf{NetworkX} \citep{SciPyProceedings_11} is an open-source Python Library for creating and studying complex networks. It is endowed with several standard graph algorithms and data structures for graphs. In this work, we use NetworkX to make statistics on graphs, namely number of nodes, number of edges, and density, as well connectivity. We have used version 2.6 of NetworkX in this work. NetworkX is released as free and open-source \emph{Berkeley Software Distribution License}.

\textbf{PyTorch} \citep{NEURIPS2019_9015} is an open-source library for DL using GPUs and CPUs. In this work, we use PyTorch to build our training protocol. In this work, we have used version 1.9.1 of Pytorch along with CUDA 11.1. PyTorch is released as free and open-source \emph{Berkeley Software Distribution License}.

\textbf{PyTorch Geometric} (PyG) \citep{fey2019graph} is an open-source library for GDL built upon PyTorch which targets the training  of geometric neural networks, including point clouds, graphs and meshes. We use PyG to design our message passing schemes. In this work, we have used version 2.0.2 of PyG along with CUDA 11.1. PyG is released as free and open-source software under the \emph{MIT License}.

\begin{figure}
    \centering
    \includegraphics[width = \linewidth]{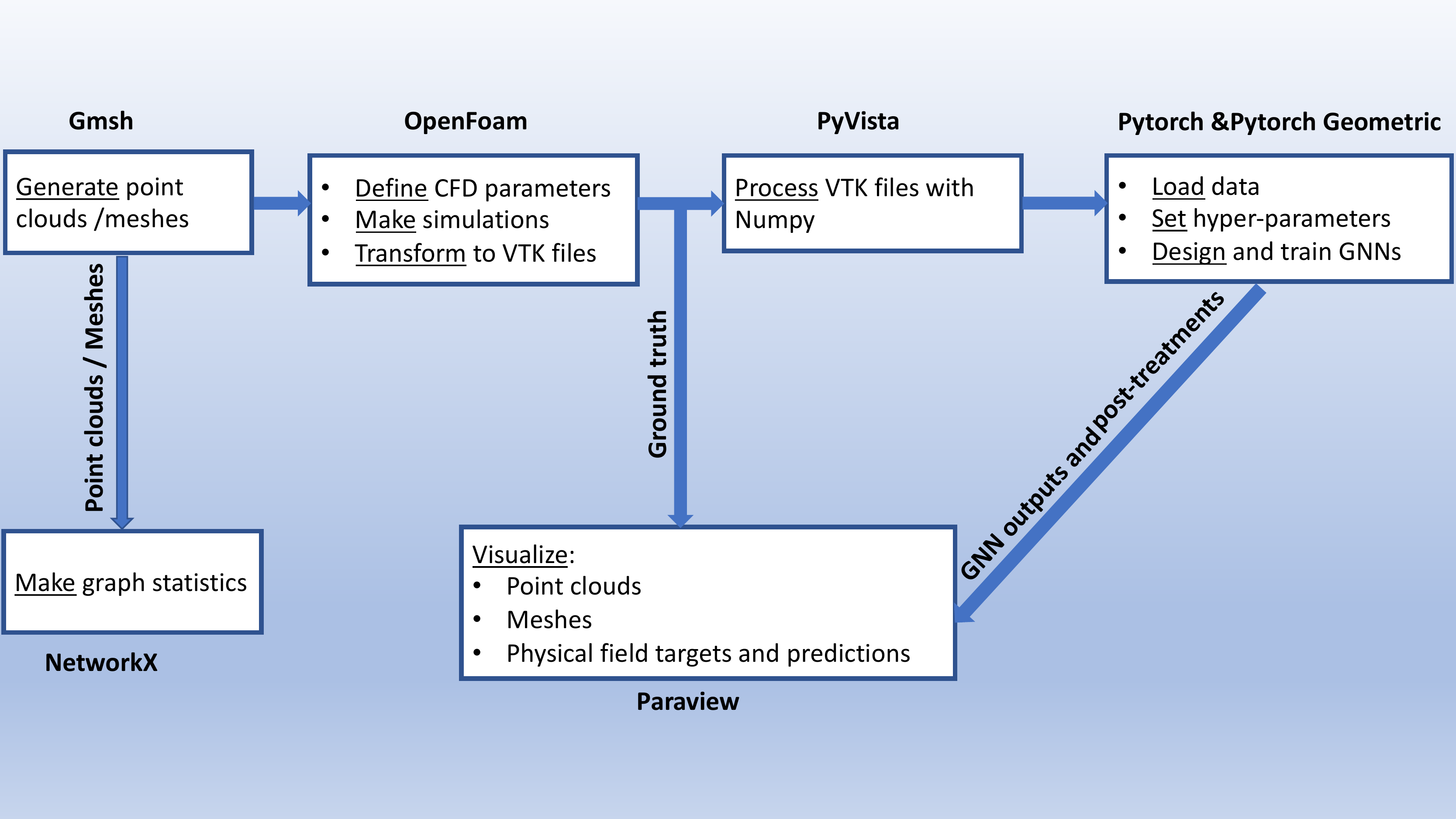}
    \caption{The pipeline to make the experiments and the connection between the different tools. CFD, VTK, GNNs stand respectively for Computational Fluid Dynamic, Visualization Toolkit, Graph Neural Networks.}
    \label{fig:working_pipline}
\end{figure}

In Figure \ref{fig:working_pipline}, we illustrate the whole pipeline to make the experiments. Starting from mesh generation to model training and output visualizations. We show the connection between all the aforementioned tools to perform our task.

\section{Graph generation: from geometries to CFD meshes and radius graphs.}\label{ap:graph_mesh_gen}

We start the CFD process with only the different airfoils as input. We load these geometries (under the form of point clouds) in Gmsh and reconstruct a spline interpolating the point clouds in order to recover continuous geometries. We specify the density of points we want at the surface of those geometries and at the boundaries of the domains. Then, we use Gmsh to create 2-D conformal meshes made of triangles and extrude them along the $z$-direction to get 3-D conformal meshes made of one tetrahedral in the $z$-direction. We need to do the extrusion step as OpenFOAM only works with 3-D meshes. An example of such mesh is given in Figure \ref{fig:dataset_sample}.

\section{Preprocessing}\label{ap:stat_preprocessing}

In this section, we describe the process of input and output variables normalization that we use prior to feeding them to DL models.
We compute the mean value $\mu_k$ and the standard deviation $\sigma_k$, of each component $k\in\{0, 1, 2, 3\}$ of all the nodes inputs of all the samples in the training set and we normalize each sample's inputs $x$ as follow:
\begin{align}
    x^{'}_k = \frac{x_k - \mu_k}{\sigma_k + 10^{-8}}
\end{align}
where the factor $10^{-8}$ is added for numerical stability. For the targets, a similar process is applied. However, we observed a particularity in the distribution of the turbulent viscosity at the surface compared to the one in the volume. The values of the turbulent viscosity at the surface are of, at least, one order of magnitude smaller. Hence, we have chosen to normalize it independently from the volume values. This trick leads to a better performance on the inference of the turbulent viscosity over the surface. The transformation applied to the targets are as follow, we first normalize the targets $y$ via:
\begin{align}
    y^{'}_k = \frac{y_k - \mu_k}{\sigma_k + 10^{-8}}
\end{align}
where $\mu_k$ is the mean value of the $k$-component of all of the targets of all the samples in the training set and $\sigma_k$ their standard deviation. In addition to that, for the turbulent viscosity component (\emph{i.e.} for $k = 4$) associated to the nodes at the surface of airfoils, we apply a second transformation as follow:
\begin{align}
    y^{''}_4 &= \frac{y^{'}_4(\sigma_4 + 10^{-8}) + \mu_4 - \mu^{(s)}_4}{\sigma^{(s)}_4 + 10^{-8}} \\
    &= \frac{y^{(s)}_4 - \mu^{(s)}_4}{\sigma^{(s)}_4 + 10^{-8}}
\end{align}
where $\mu^{s}_4$ is the mean value of the turbulent viscosity of all surface nodes of all samples in the training set and $\sigma^{s}_4$ their standard deviation. These mean and standard deviation values are used to normalize the validation set and test set.

\section{Dataset figures}\label{sec:dataset_visualization}
\begin{figure}[ht]
    \begin{subfigure}{\textwidth}
        \centering
        \includegraphics[width = 0.8\textwidth]{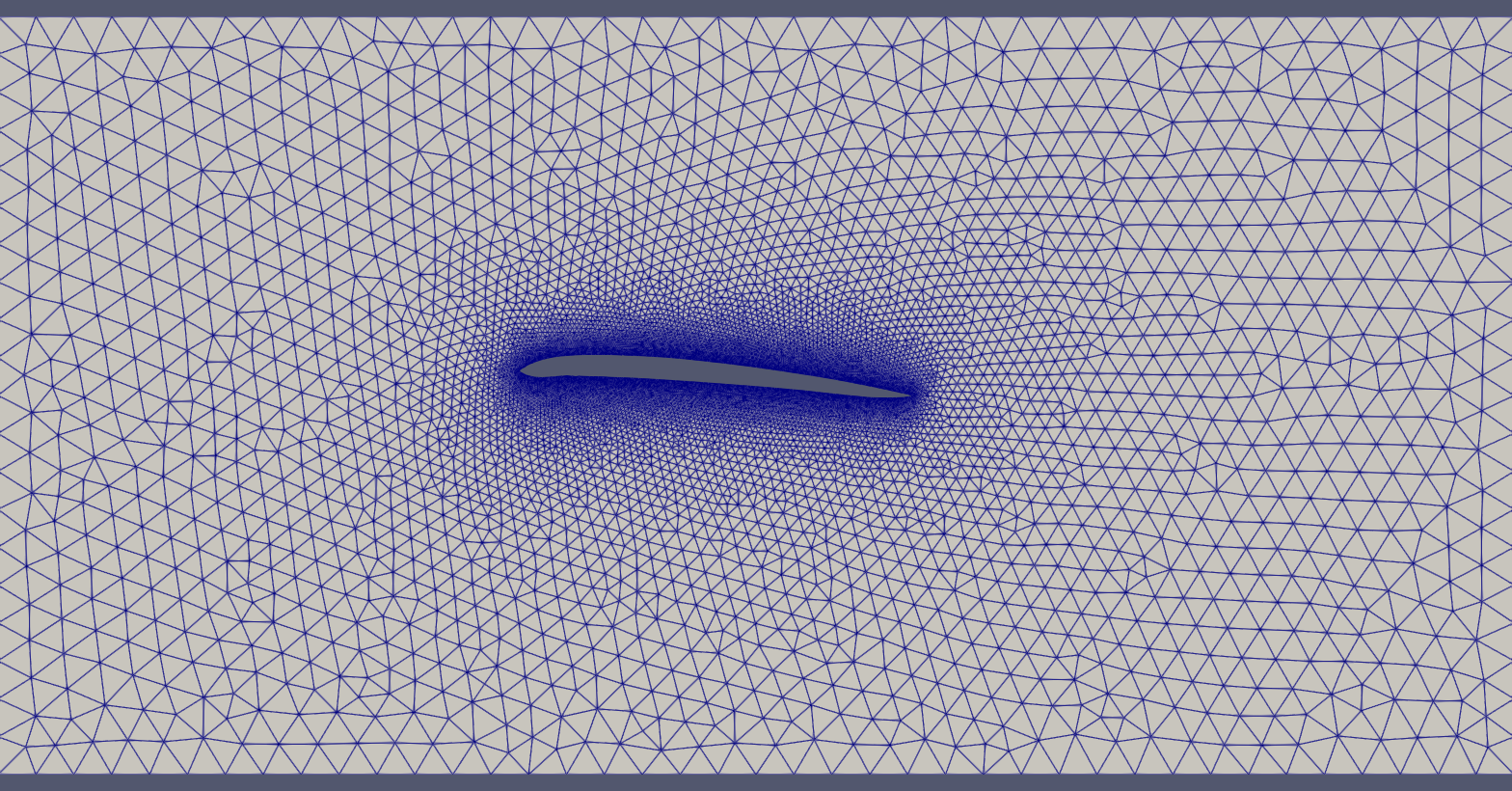}  
        \caption{Example in the training set.}
    \end{subfigure}
    \begin{subfigure}{\textwidth}
        \centering
        \includegraphics[width = 0.8\textwidth]{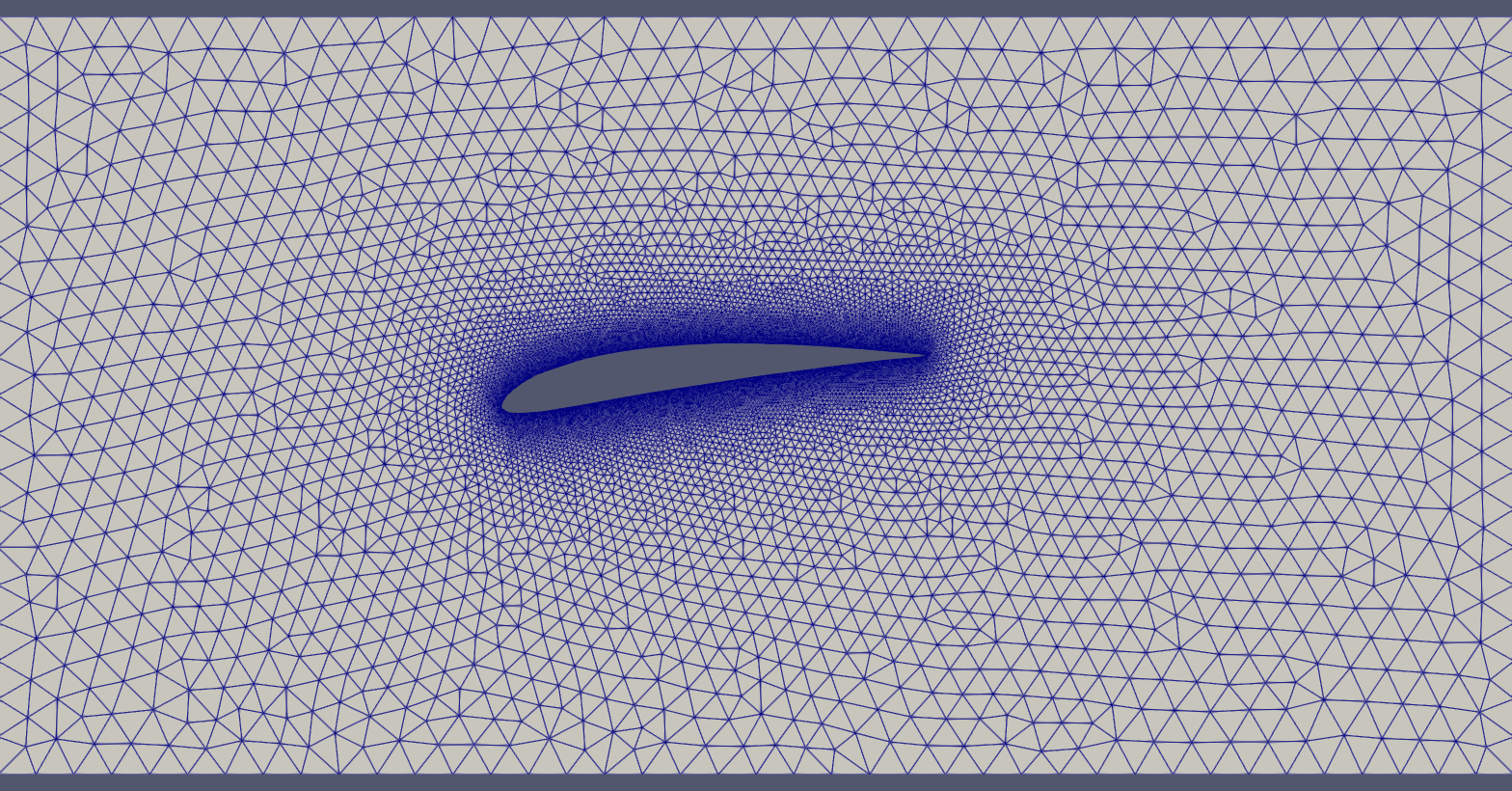}  
        \caption{Example in the validation set.}
    \end{subfigure}
    \begin{subfigure}{\textwidth}
        \centering
        \includegraphics[width = 0.8\textwidth]{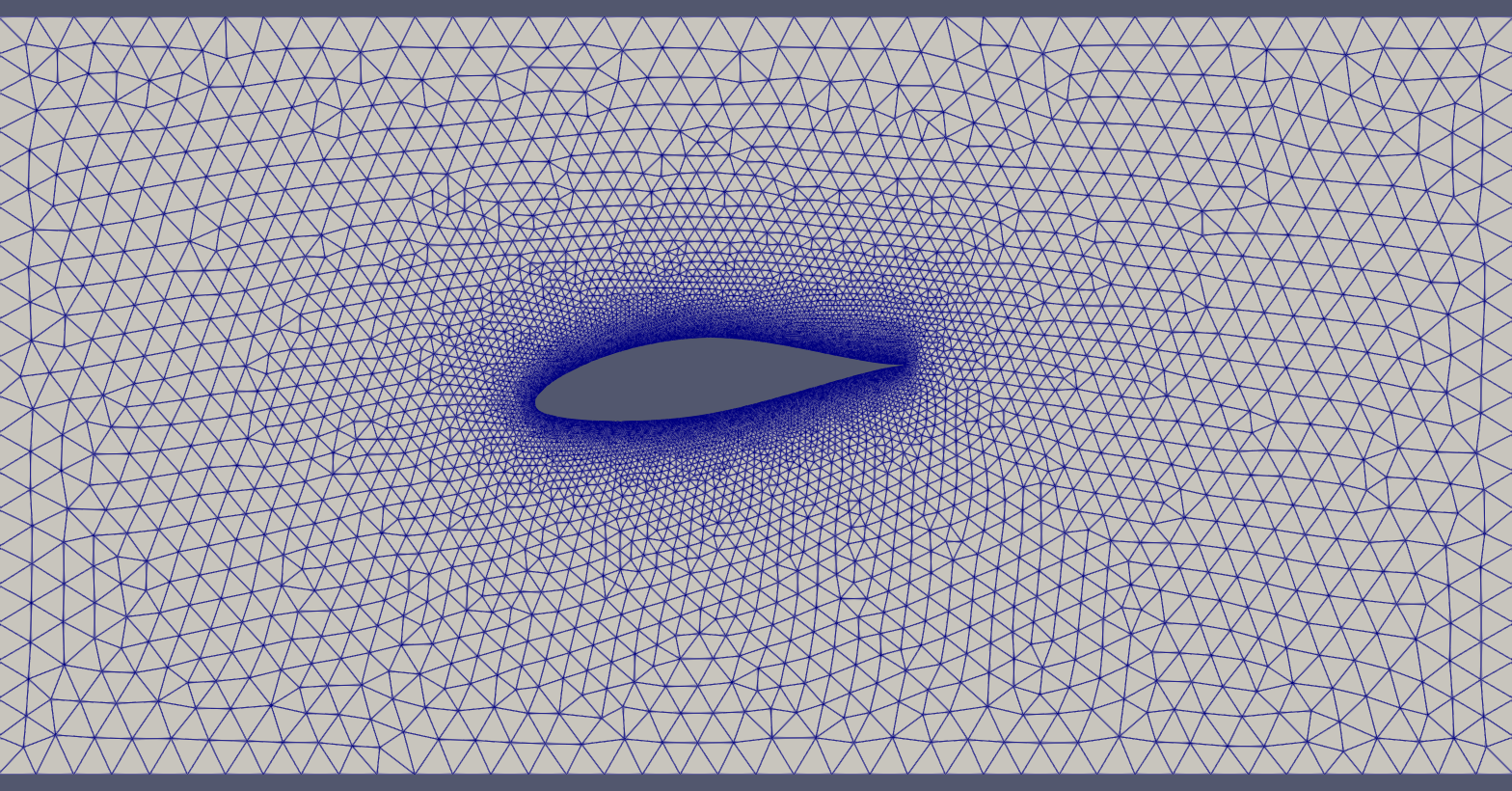}  
        \caption{Example in the test set.}
    \end{subfigure}
    \caption{Examples of CFD meshes in the dataset.}
    \label{fig:dataset_sample}
\end{figure}

\begin{figure}[ht]
    \begin{subfigure}{\textwidth}
        \centering

        \includegraphics[width = 0.8\textwidth]{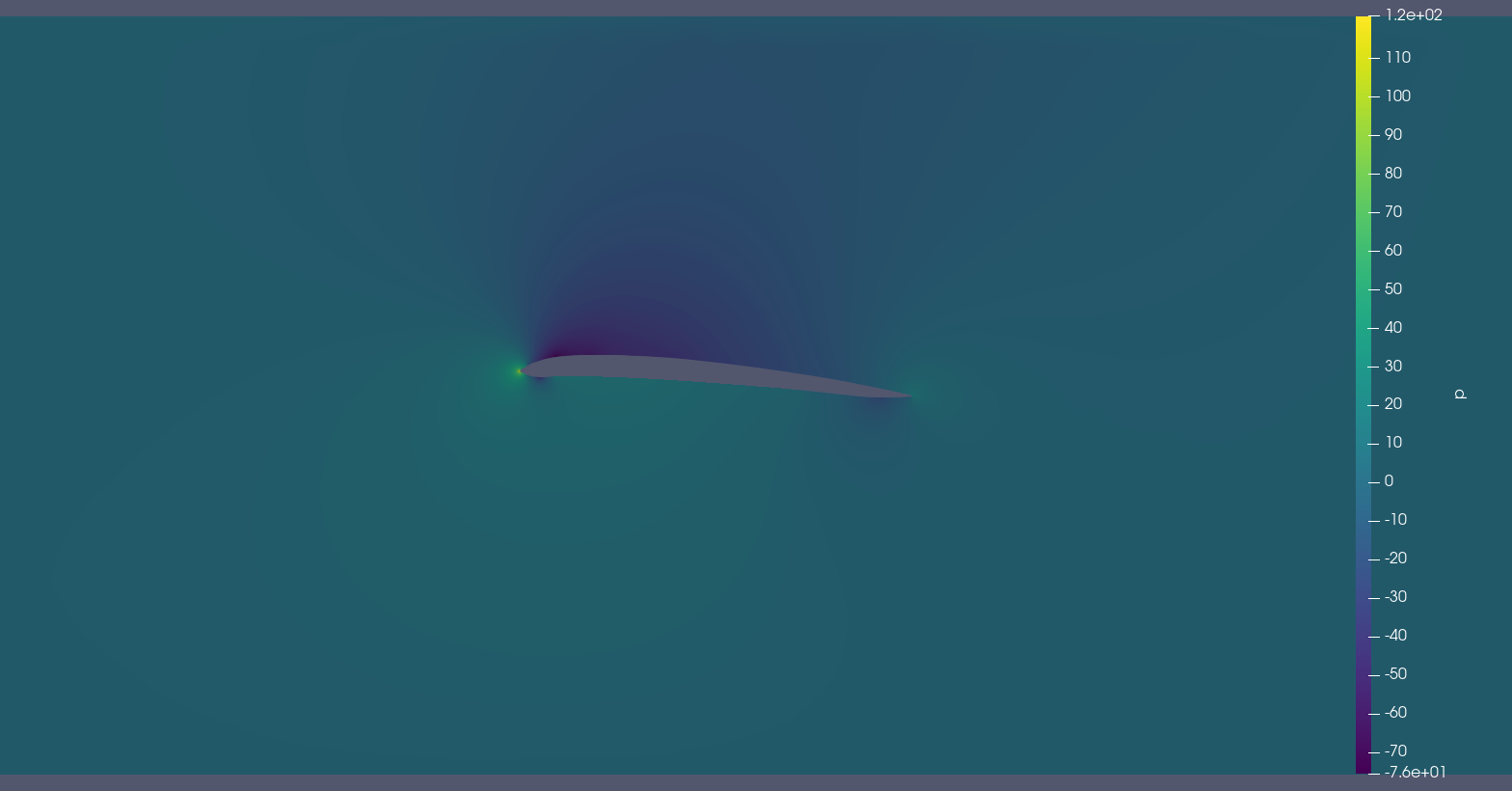}  
                        \caption{Example in the training set.}
    \end{subfigure}
    \begin{subfigure}{\textwidth}
        \centering
 
        \includegraphics[width = 0.8\textwidth]{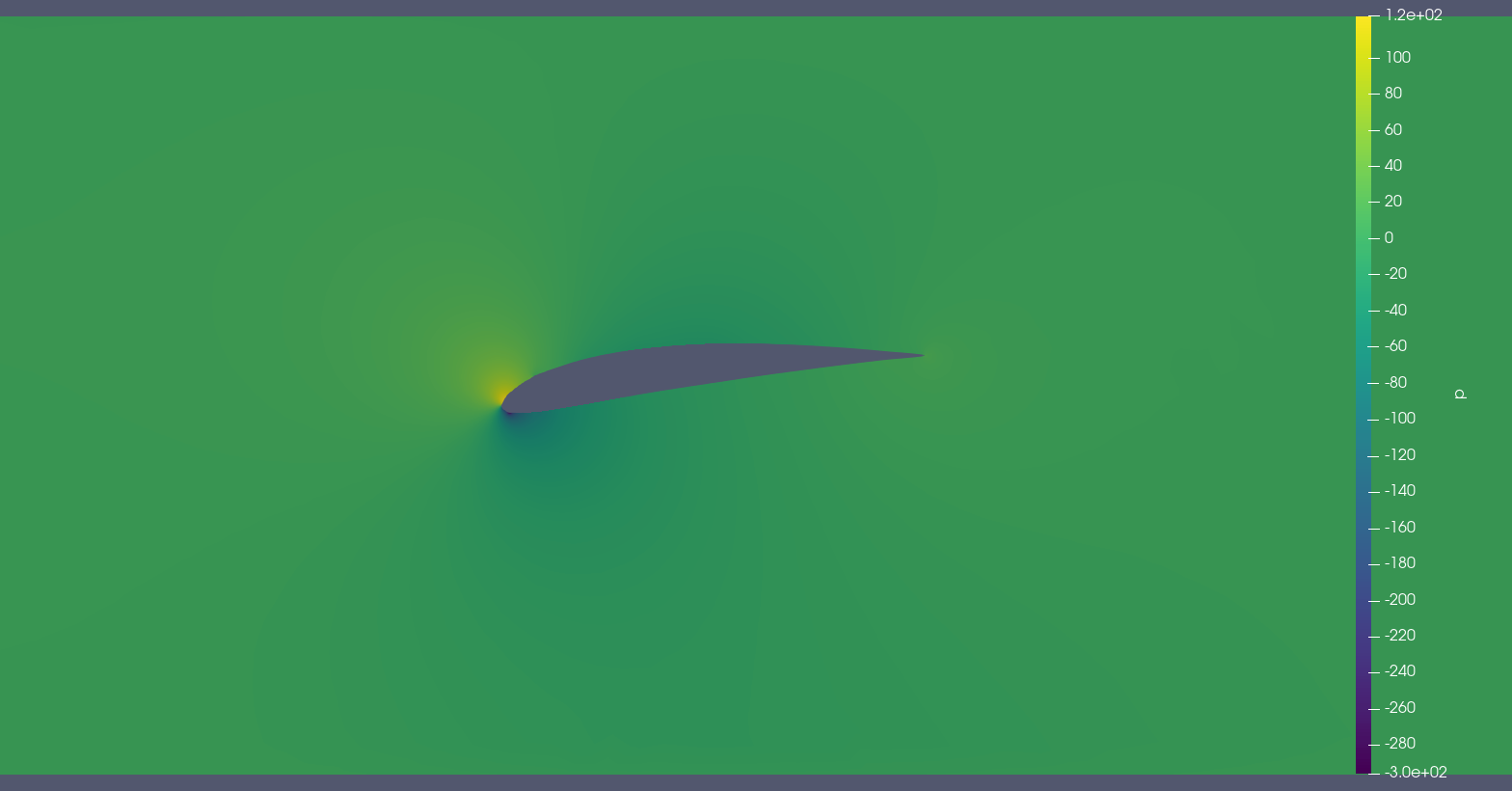}  
                       \caption{Example in the validation set.}
    \end{subfigure}
    \begin{subfigure}{\textwidth}
        \centering

        \includegraphics[width = 0.8\textwidth]{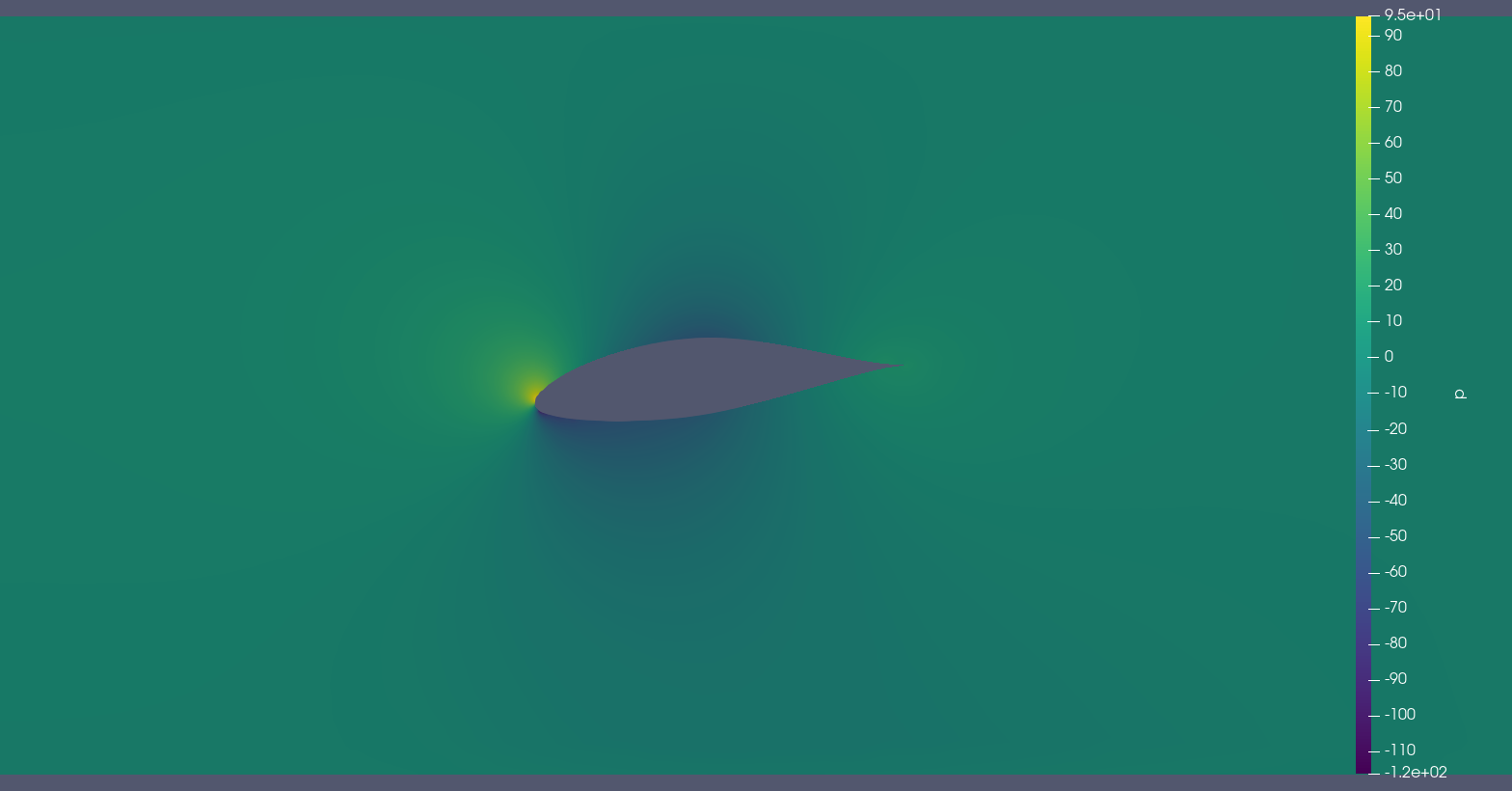}  
                \caption{Example in the test set.}
    \end{subfigure}
    \caption{Examples of pressure fields in the dataset.}
    \label{fig:dataset_pressure}
\end{figure}

In Figure \ref{fig:dataset_sample} are displayed the CFD meshes of samples from the training, validation and test sets. In Figure \ref{fig:dataset_pressure} are displayed the simulated pressure field of the same samples from the training, validation and test sets. It would be better to visualize them using Paraview for more flexibility. We release the code for that as already mentioned in Appendix \ref{sec:repro}.

\section{Details about experiments}\label{sec:details_exp}
First of all, each type of model used is preceded by an encoder and followed by a decoder. Those encoder and decoder have the same architecture for each model tested, we chose a MLP with $4-64-64-8$ neurons for the encoder and a MLP with $8-64-64-4$ neurons for the decoder. Both with ReLU activation function. Moreover, for each model tested, the encoder and decoder are trained from scratch together with the new model between. In order to have a standard deviation for the scores, we trained 10 times each model and did the statistics of the obtained results. All the models are trained using Pytorch Geometric on a single GPU (nVidia Tesla P100, 16 Go).

\begin{figure}[ht]
    \centering
    \begin{subfigure}{0.45\textwidth}
      \centering
      \includegraphics[width=\textwidth]{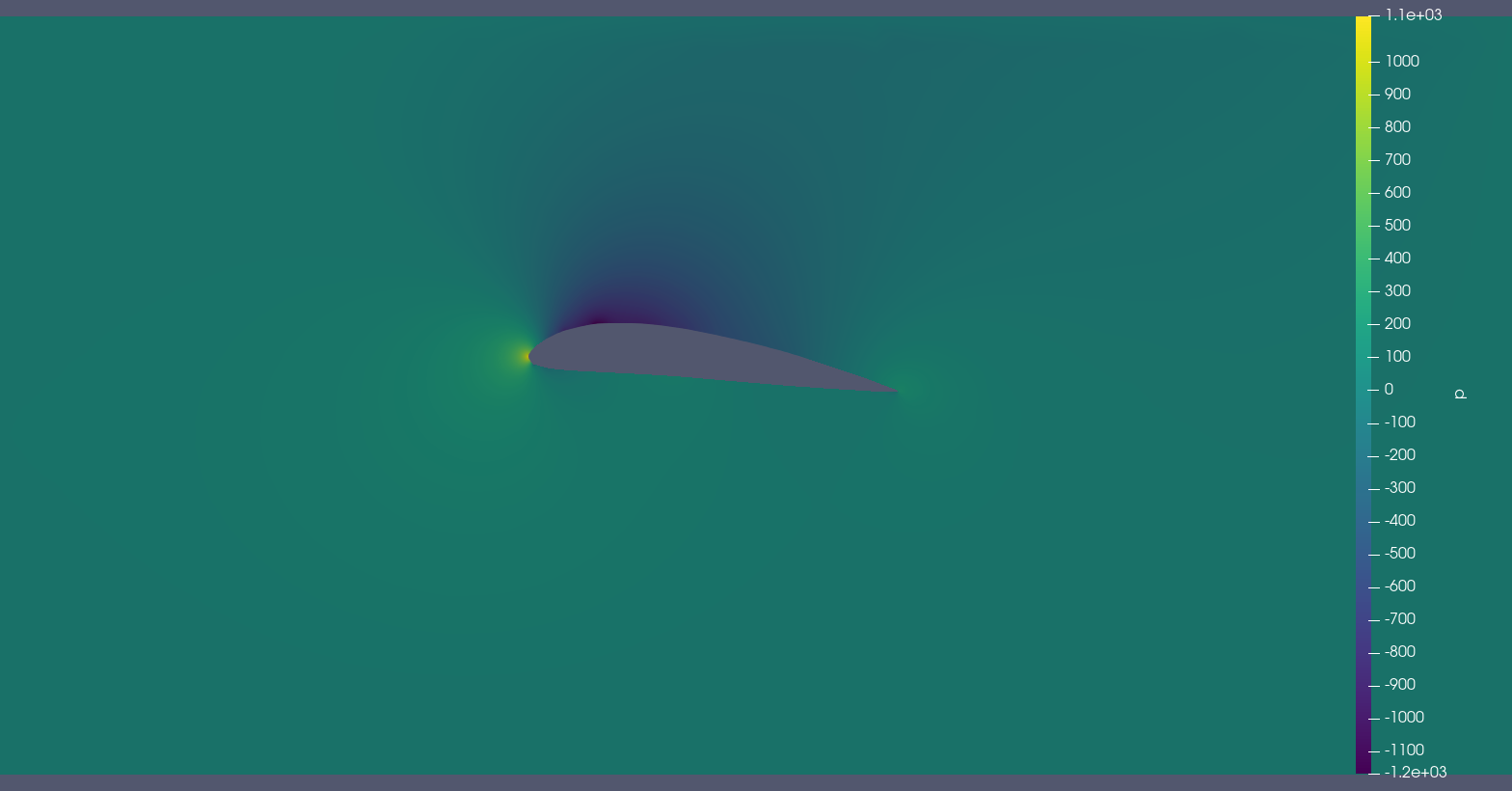}  
            \caption{Ground truth}
    \end{subfigure}
    \begin{subfigure}{0.45\textwidth}
      \centering
      \includegraphics[width=\textwidth]{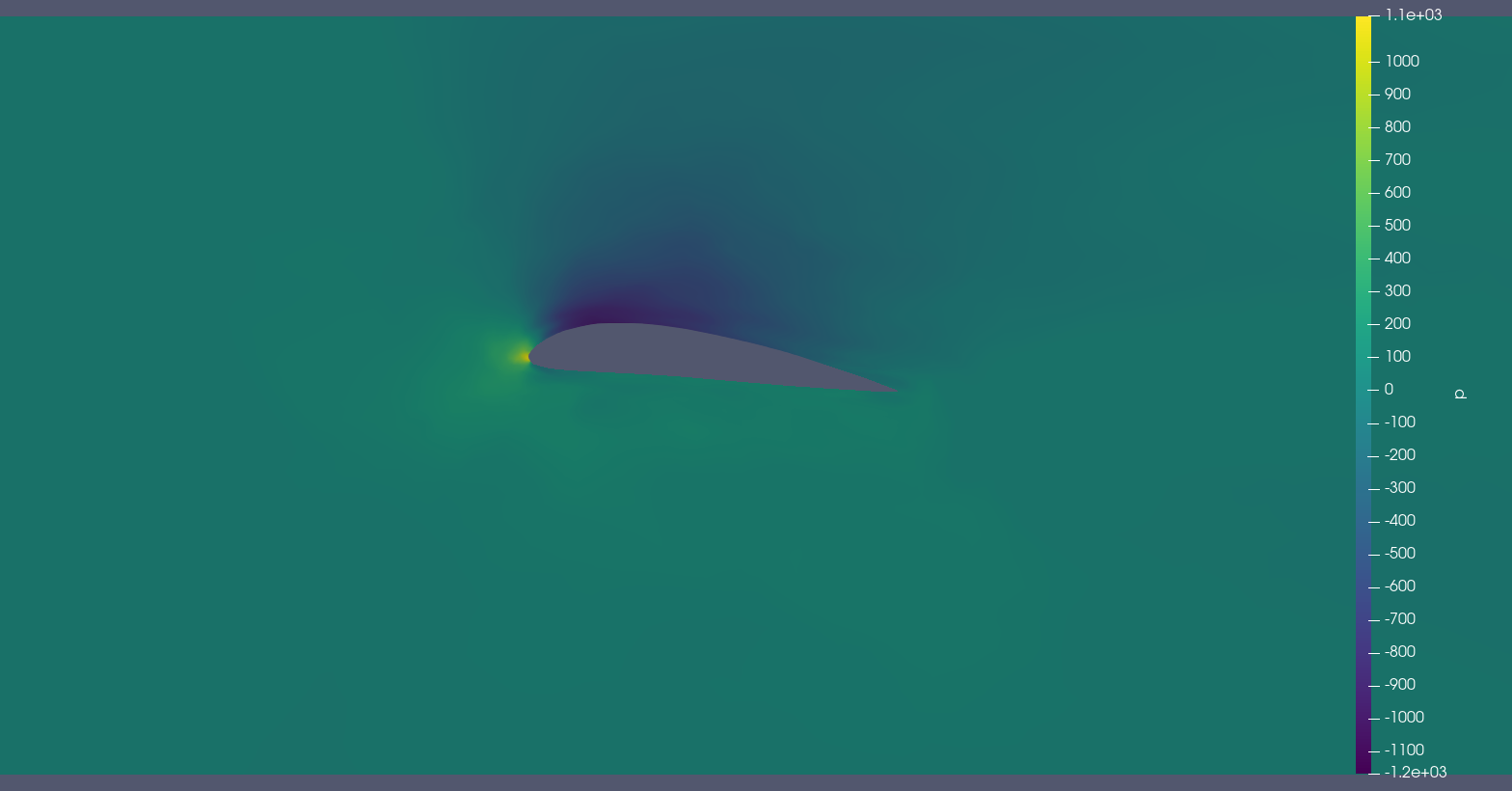}  
            \caption{GraphSAGE}
    \end{subfigure}
    \begin{subfigure}{0.45\textwidth}
      \centering
      \includegraphics[width=\textwidth]{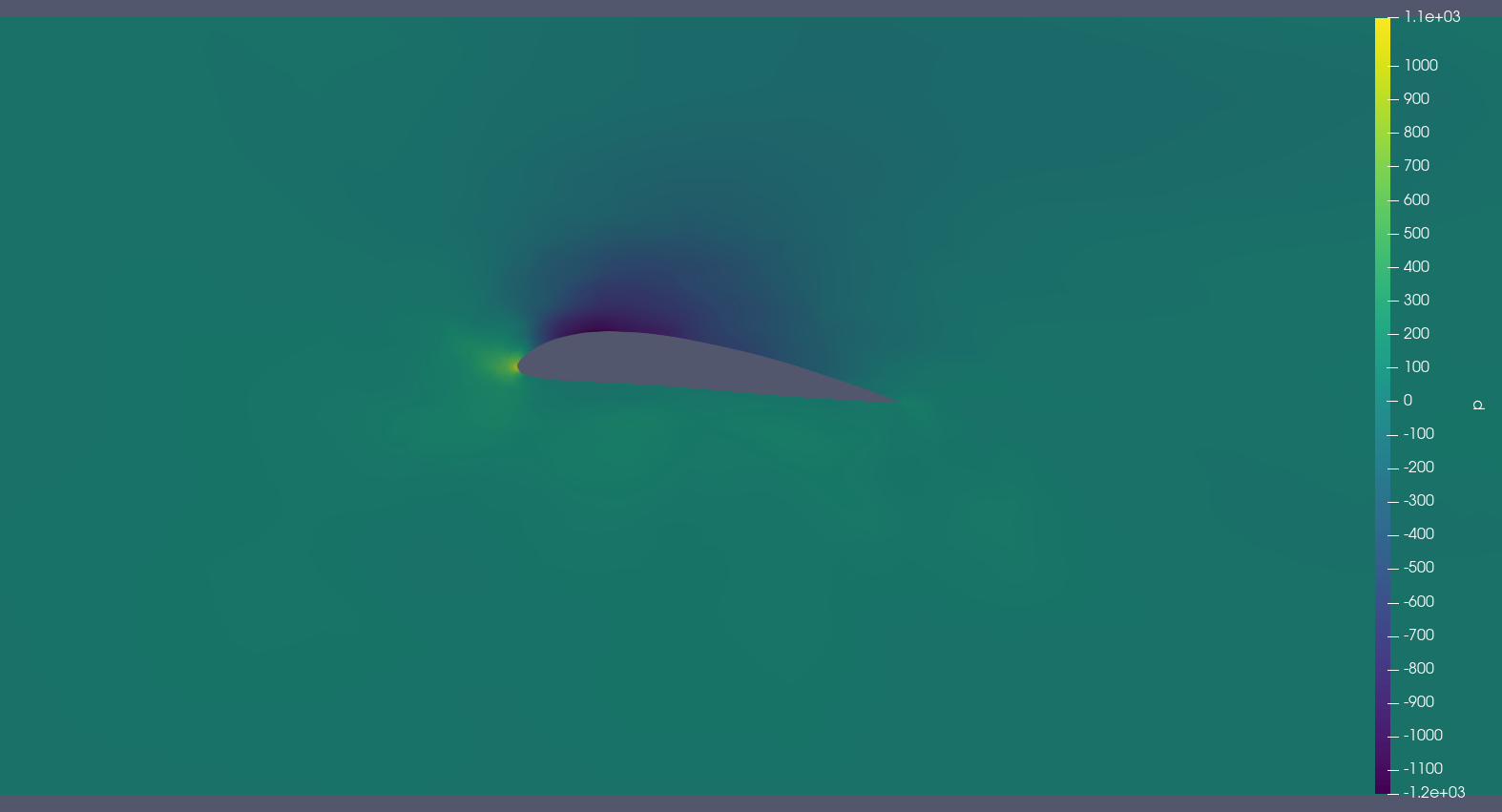}  
            \caption{GAT}
    \end{subfigure}
    \begin{subfigure}{0.45\textwidth}
      \centering
      \includegraphics[width=\textwidth]{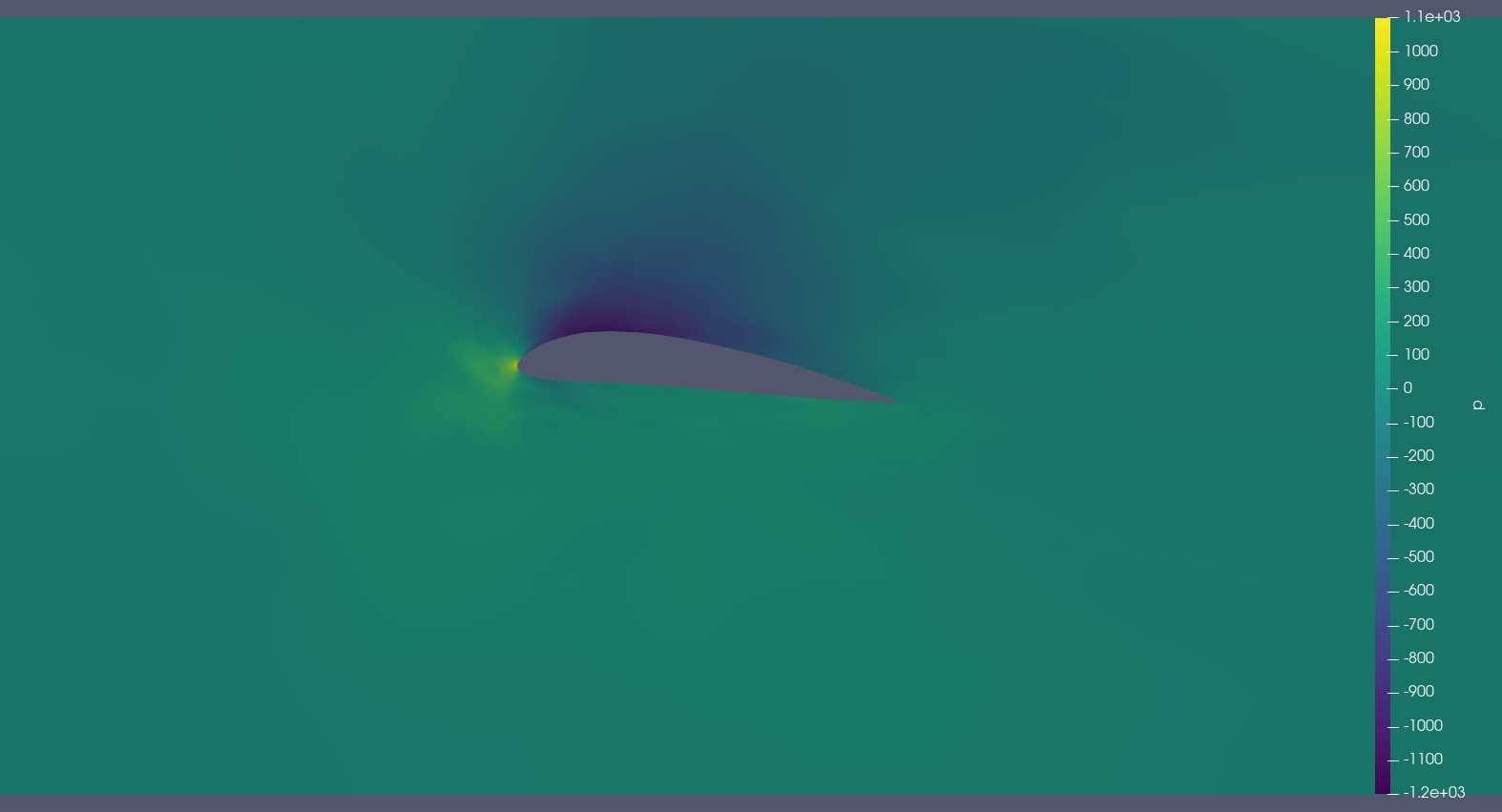}  
            \caption{PointNet}
    \end{subfigure}
    \begin{subfigure}{0.45\textwidth}
      \centering
      \includegraphics[width=\textwidth]{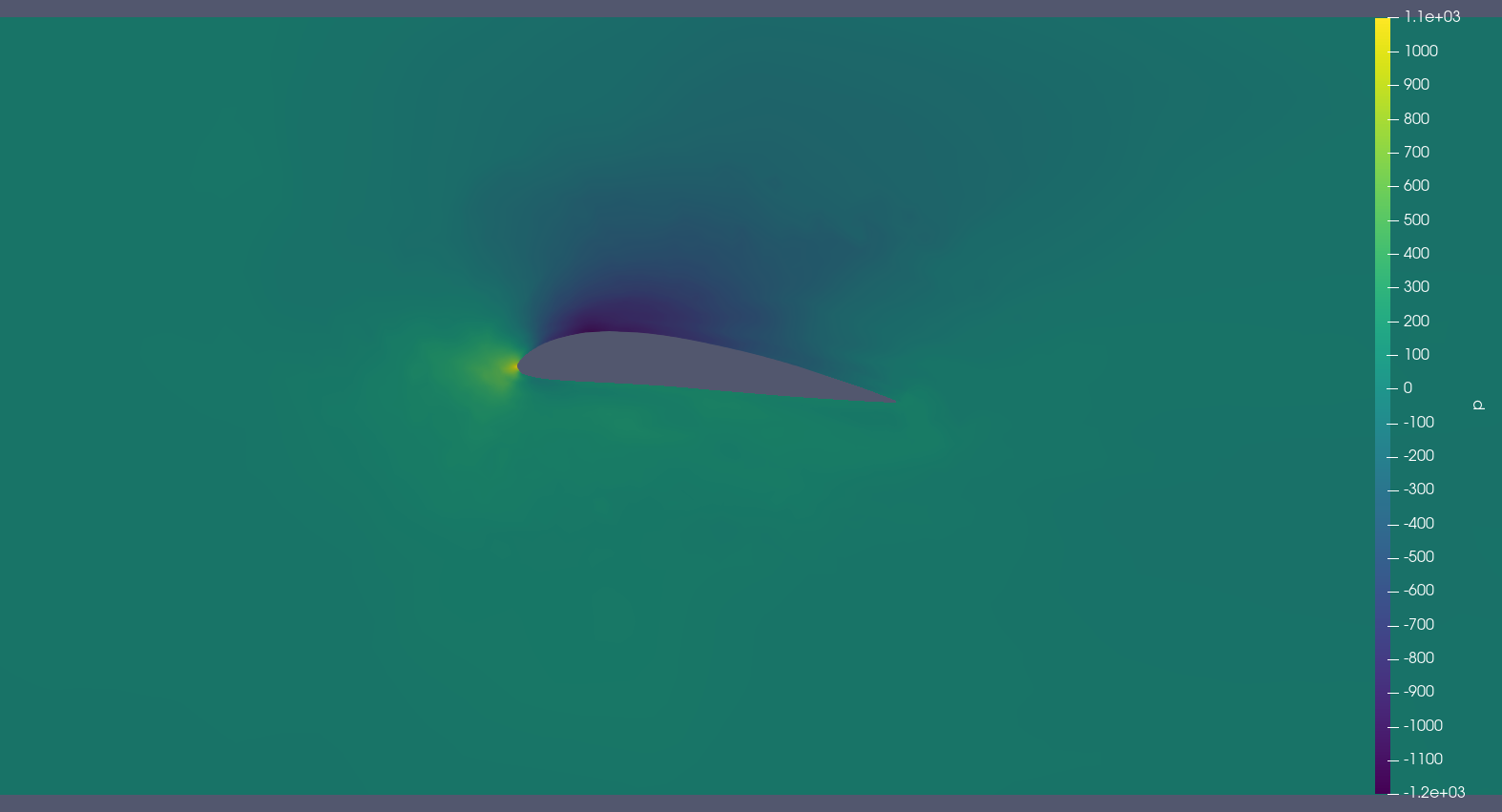}  
            \caption{GNO}
    \end{subfigure}
    \begin{subfigure}{0.45\textwidth}
      \centering
      \includegraphics[width=\textwidth]{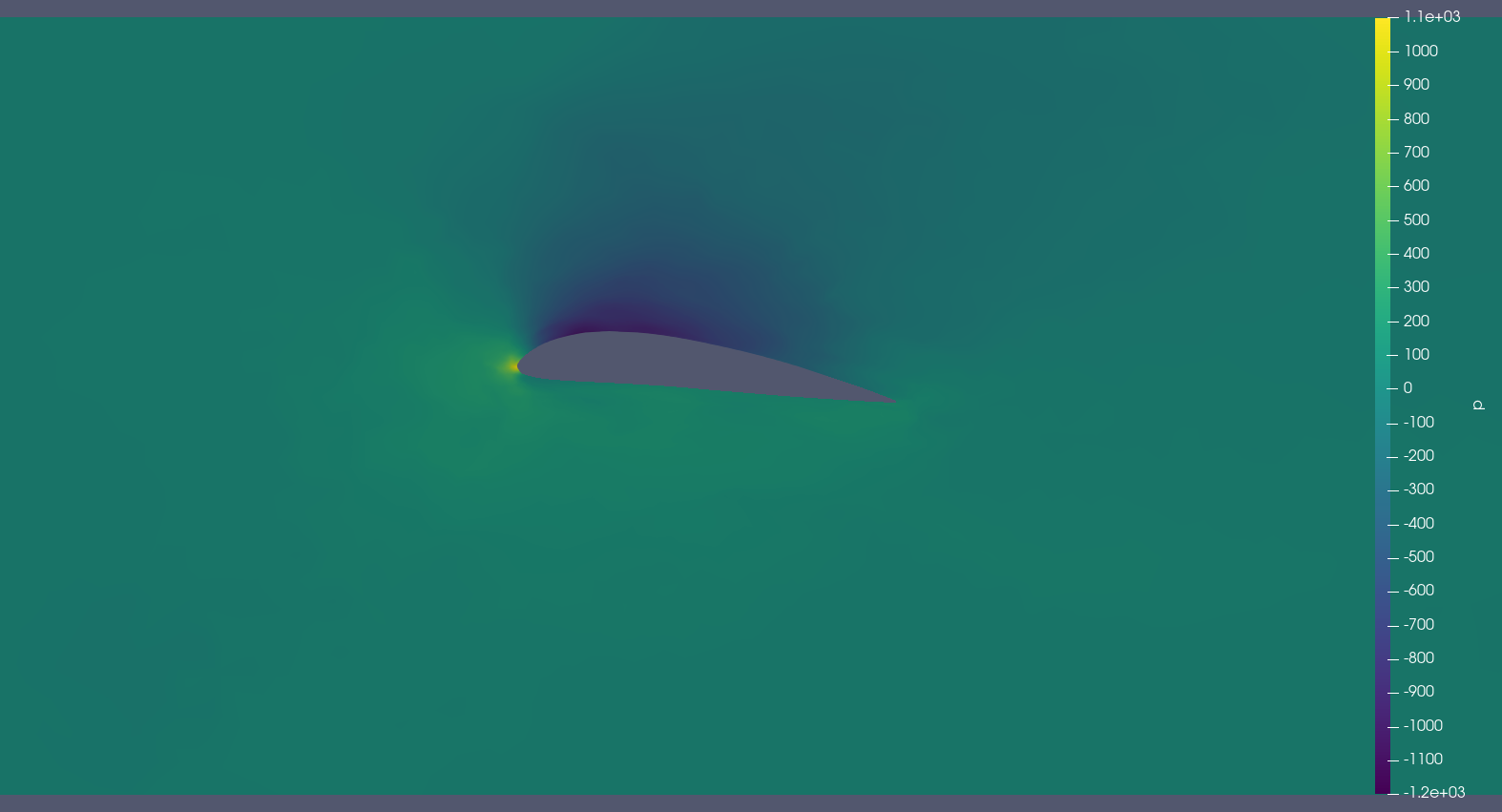}  
            \caption{Graph U-Net}
    \end{subfigure}
    \begin{subfigure}{0.45\textwidth}
      \centering
      \includegraphics[width=\textwidth]{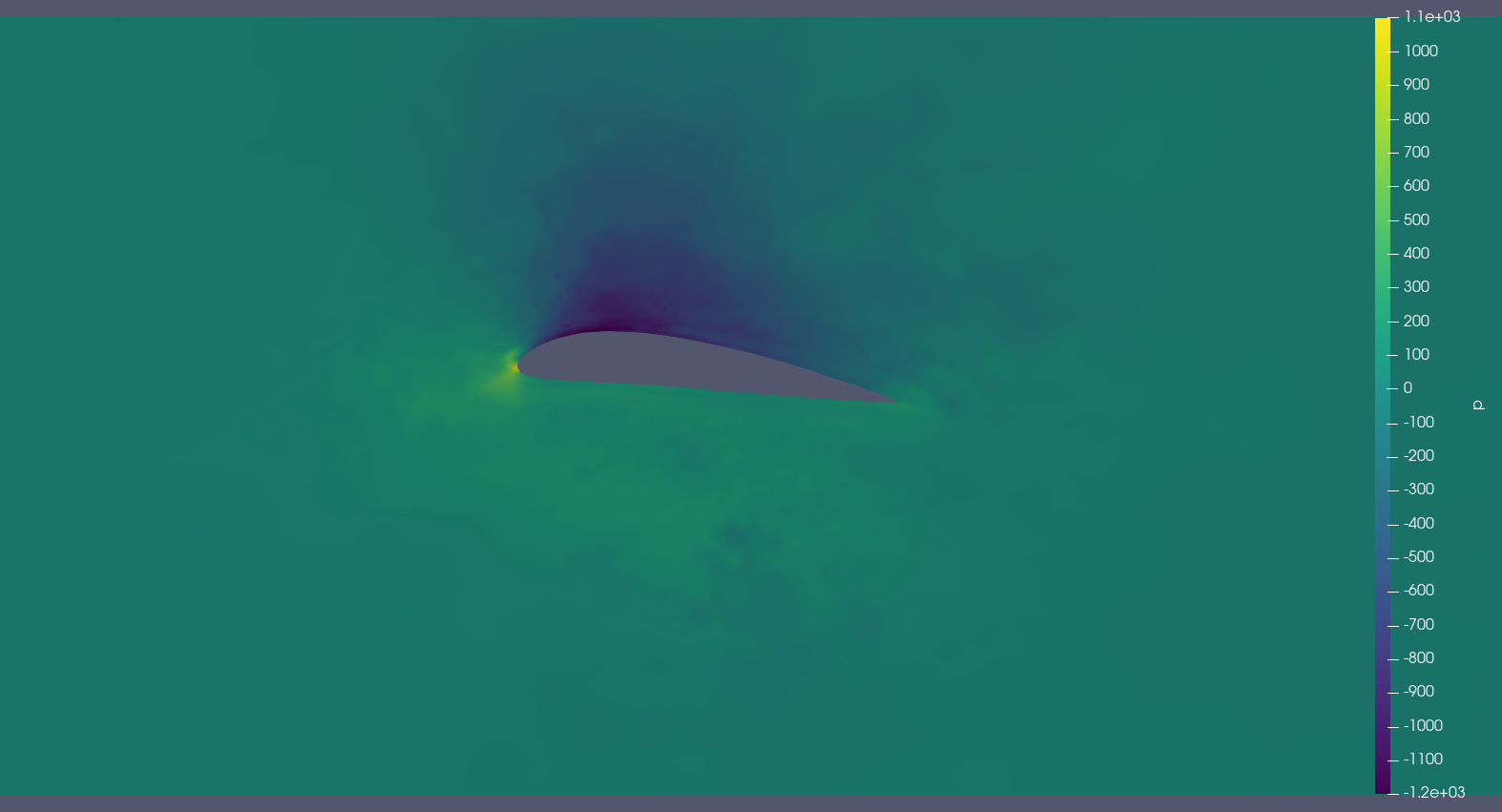}  
            \caption{PointNet++}
    \end{subfigure}
    \begin{subfigure}{0.45\textwidth}
      \centering
      \includegraphics[width=\textwidth]{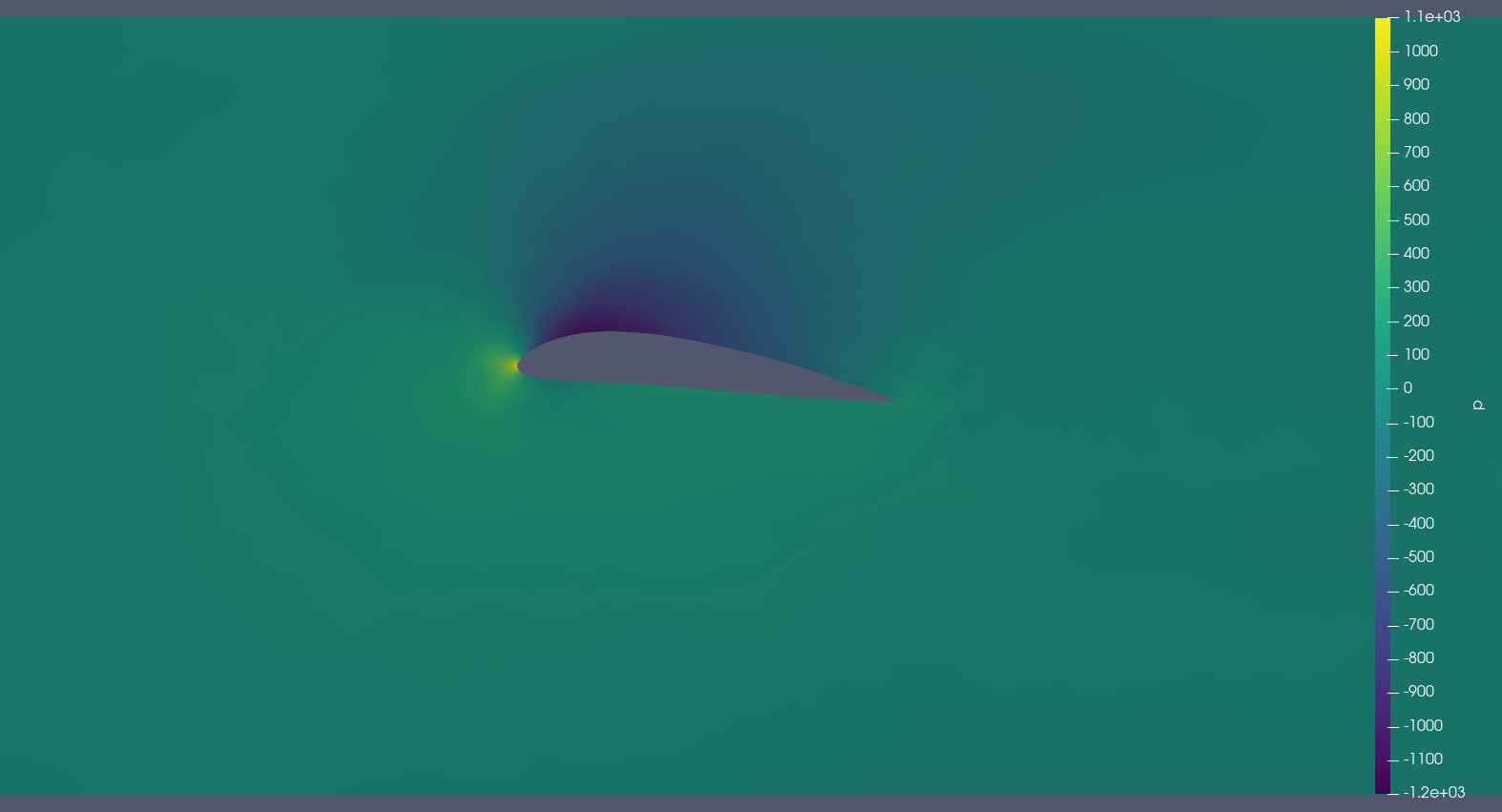}  
            \caption{MGNO}
    \end{subfigure}
    
    \caption{Comparison of the pressure of the velocity field for the different models.}
    \label{fig:qualitative_p}
\end{figure}

\begin{figure}[ht]
    \centering
    \begin{subfigure}{0.45\textwidth}
      \centering
      \includegraphics[width=\textwidth]{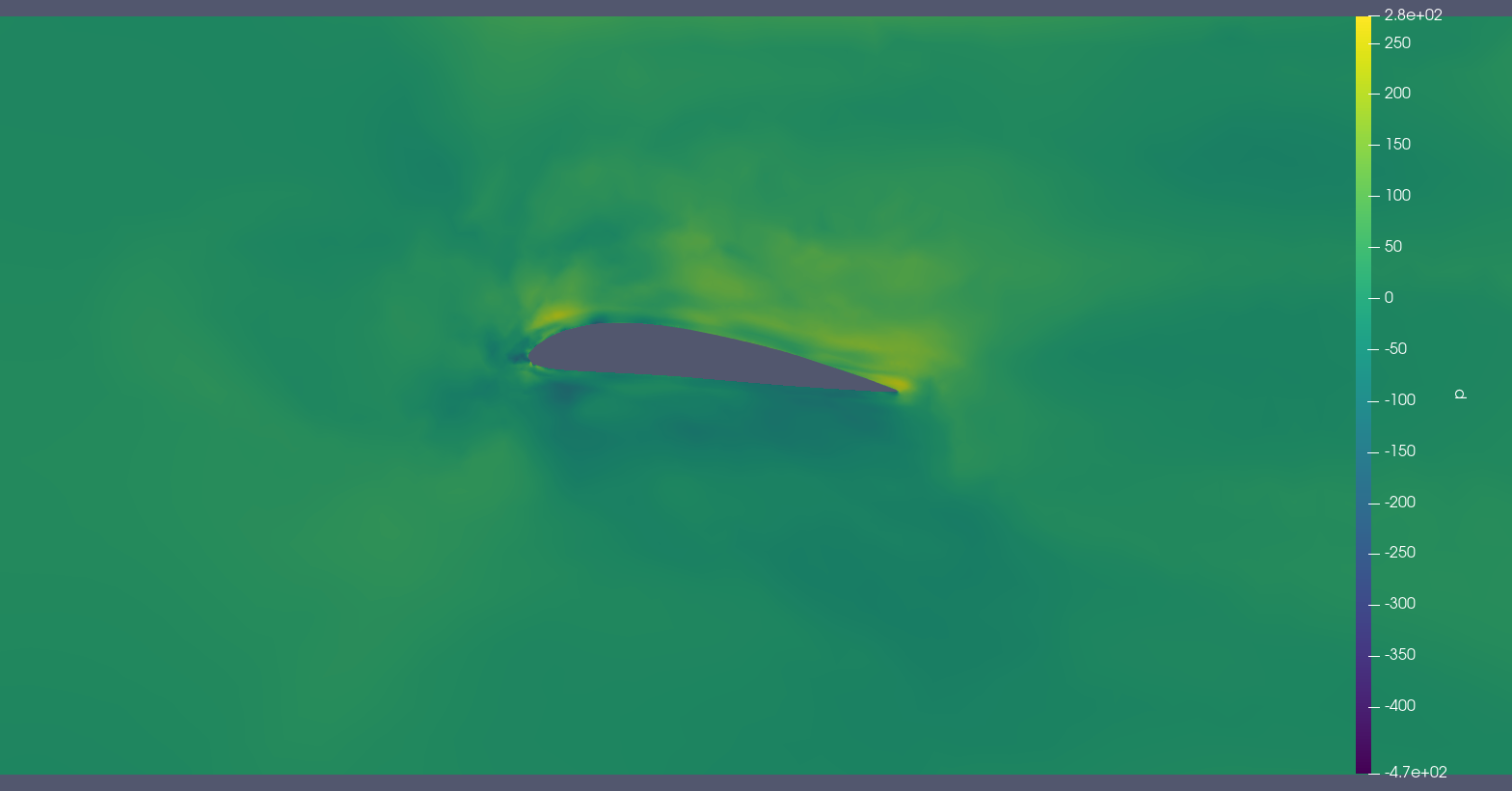}  
            \caption{GraphSAGE}
    \end{subfigure}
    \begin{subfigure}{0.45\textwidth}
      \centering
      \includegraphics[width=\textwidth]{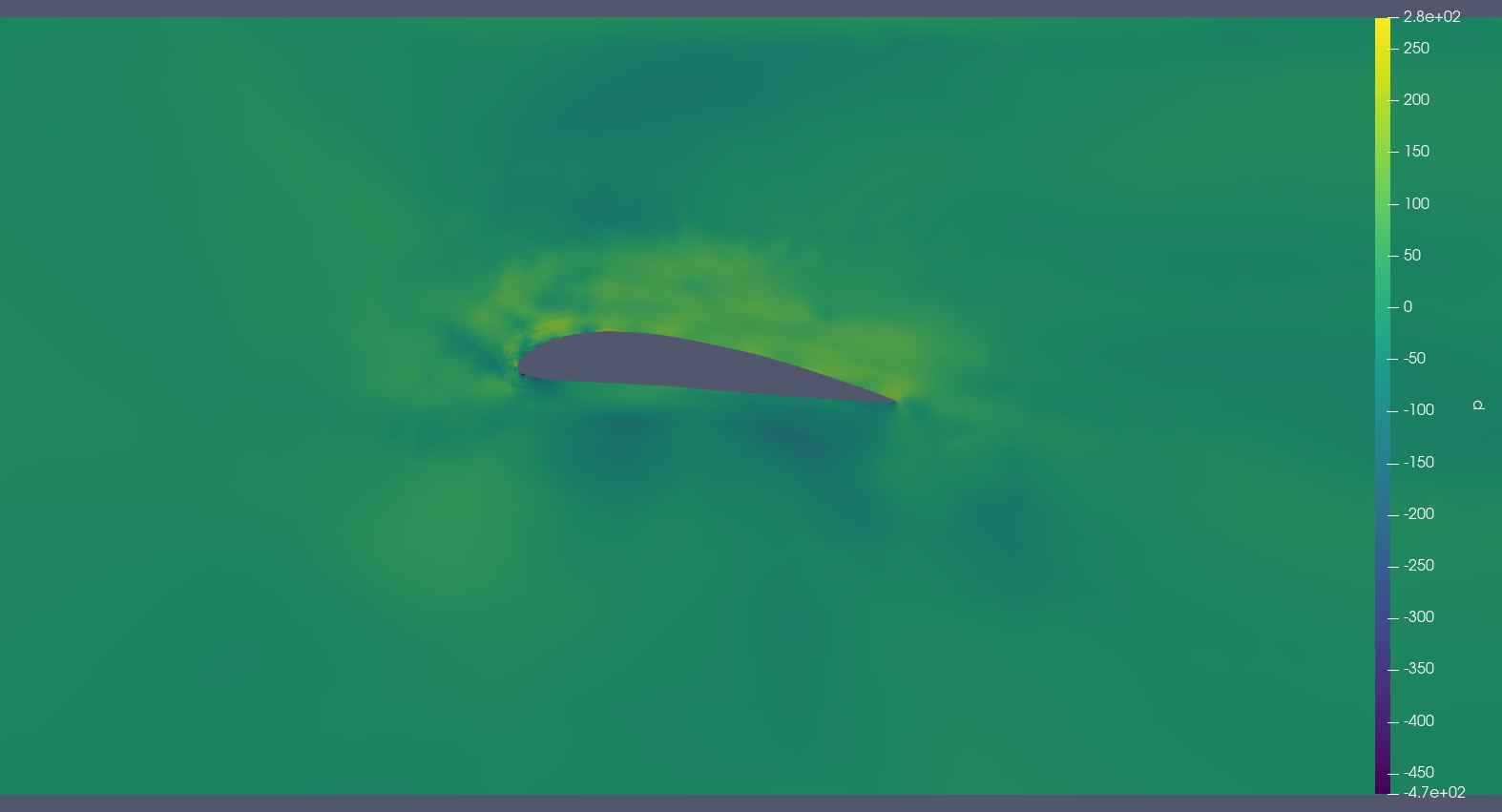}  
            \caption{GAT}
    \end{subfigure}
    \begin{subfigure}{0.45\textwidth}
      \centering
      \includegraphics[width=\textwidth]{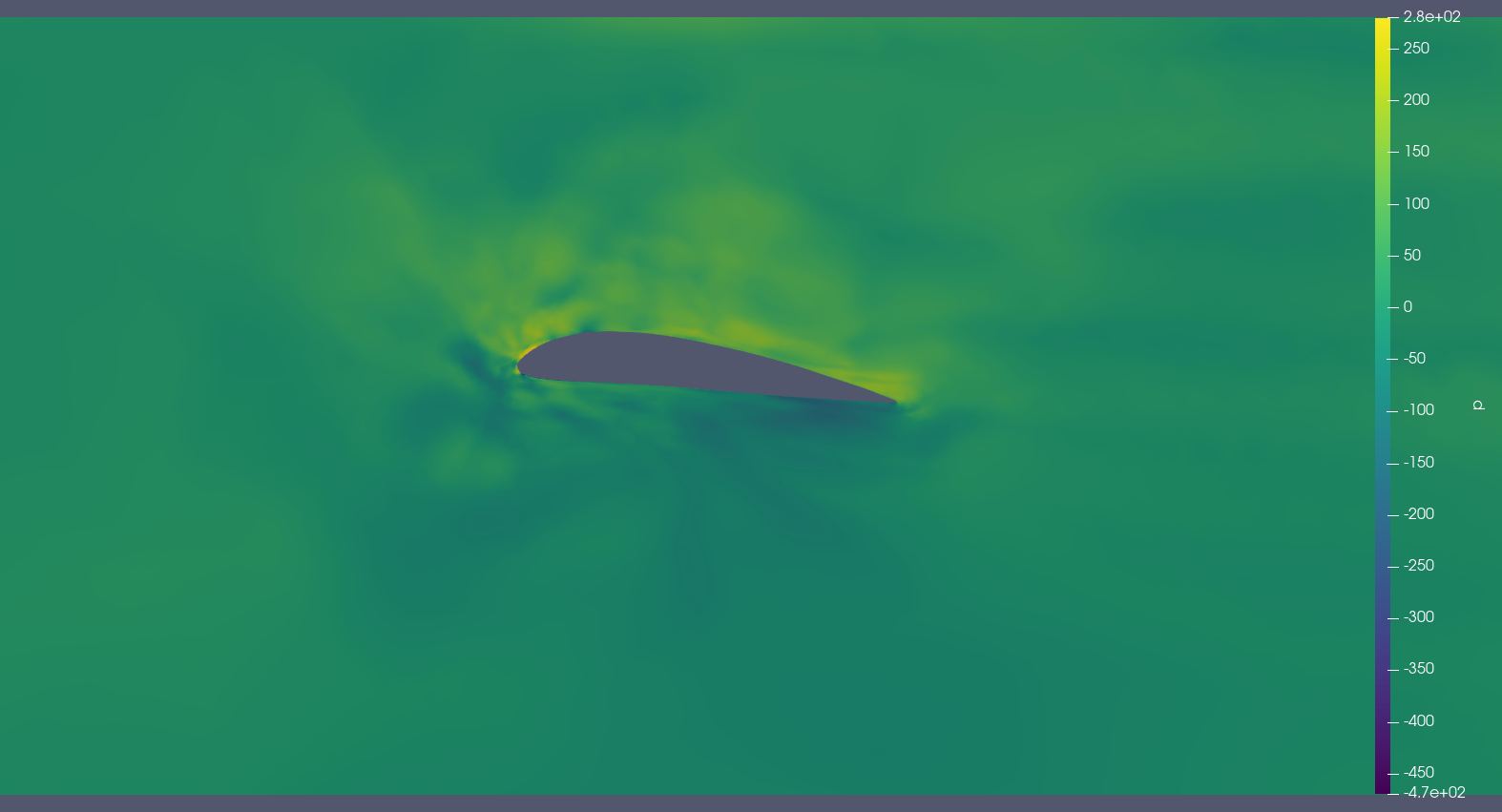}  
            \caption{PointNet}
    \end{subfigure}
    \begin{subfigure}{0.45\textwidth}
      \centering
      \includegraphics[width=\textwidth]{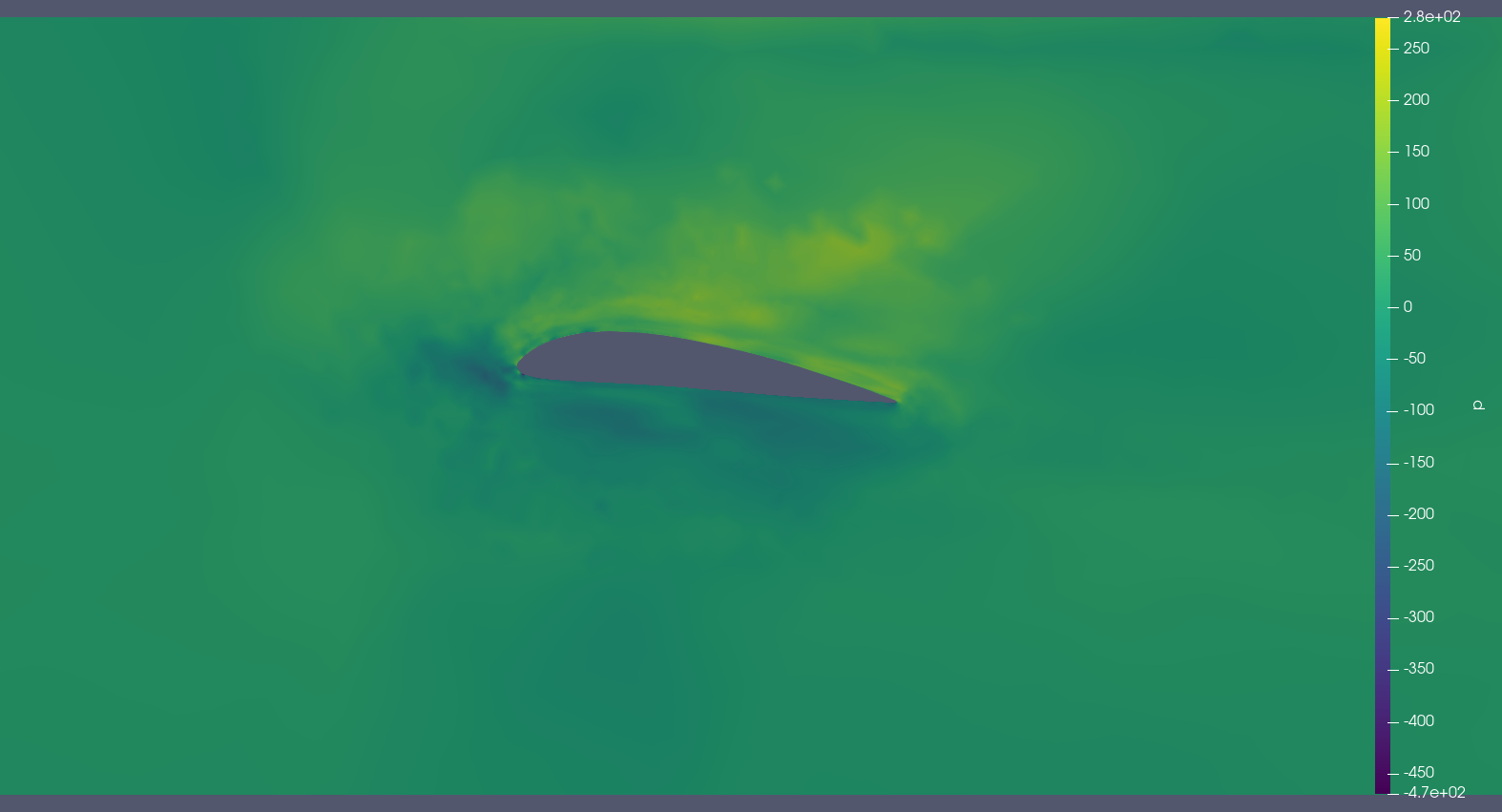}  
            \caption{GNO}
    \end{subfigure}
    \begin{subfigure}{0.45\textwidth}
      \centering
      \includegraphics[width=\textwidth]{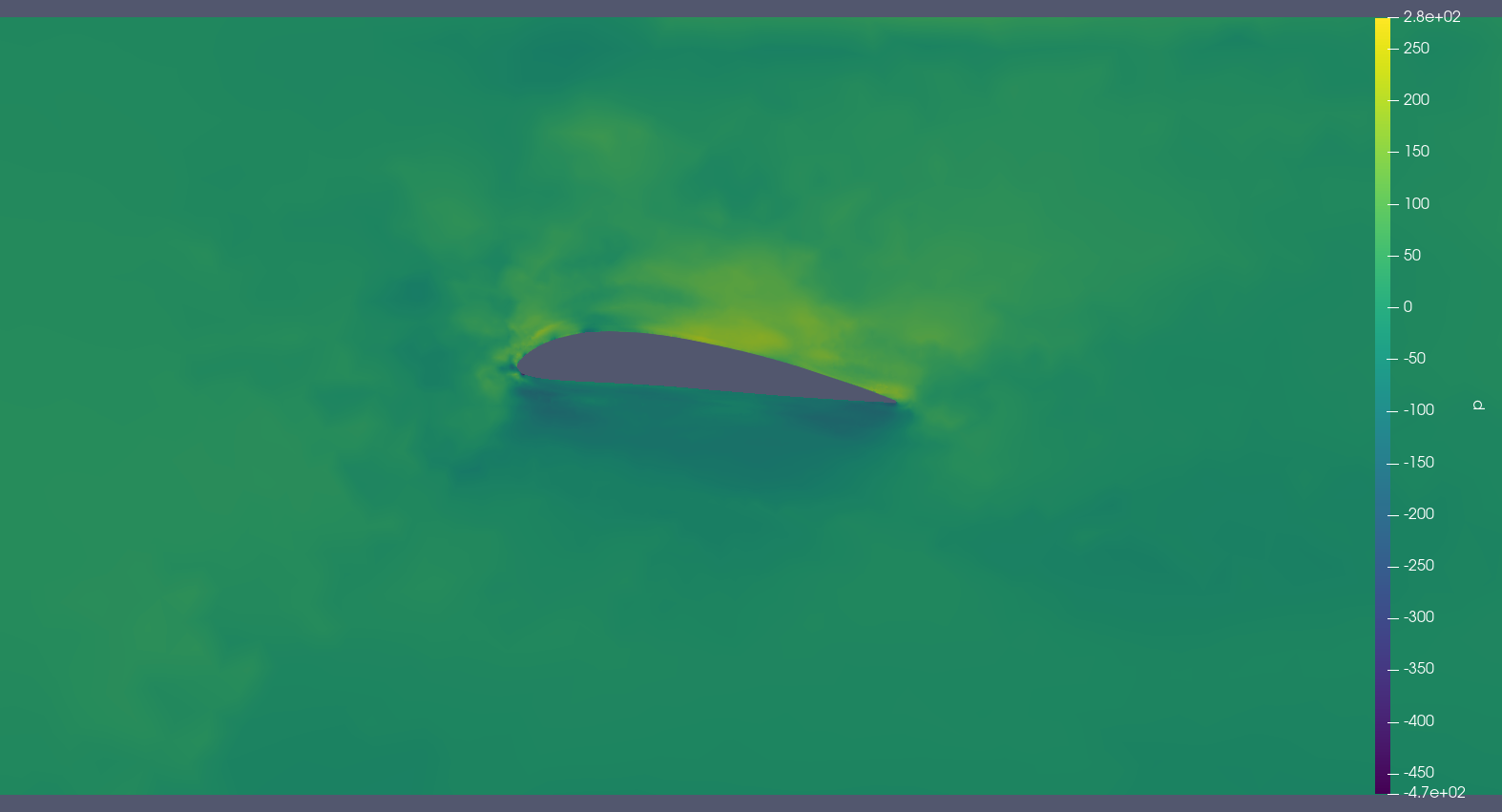}  
            \caption{Graph U-Net}
    \end{subfigure}
    \begin{subfigure}{0.45\textwidth}
      \centering
      \includegraphics[width=\textwidth]{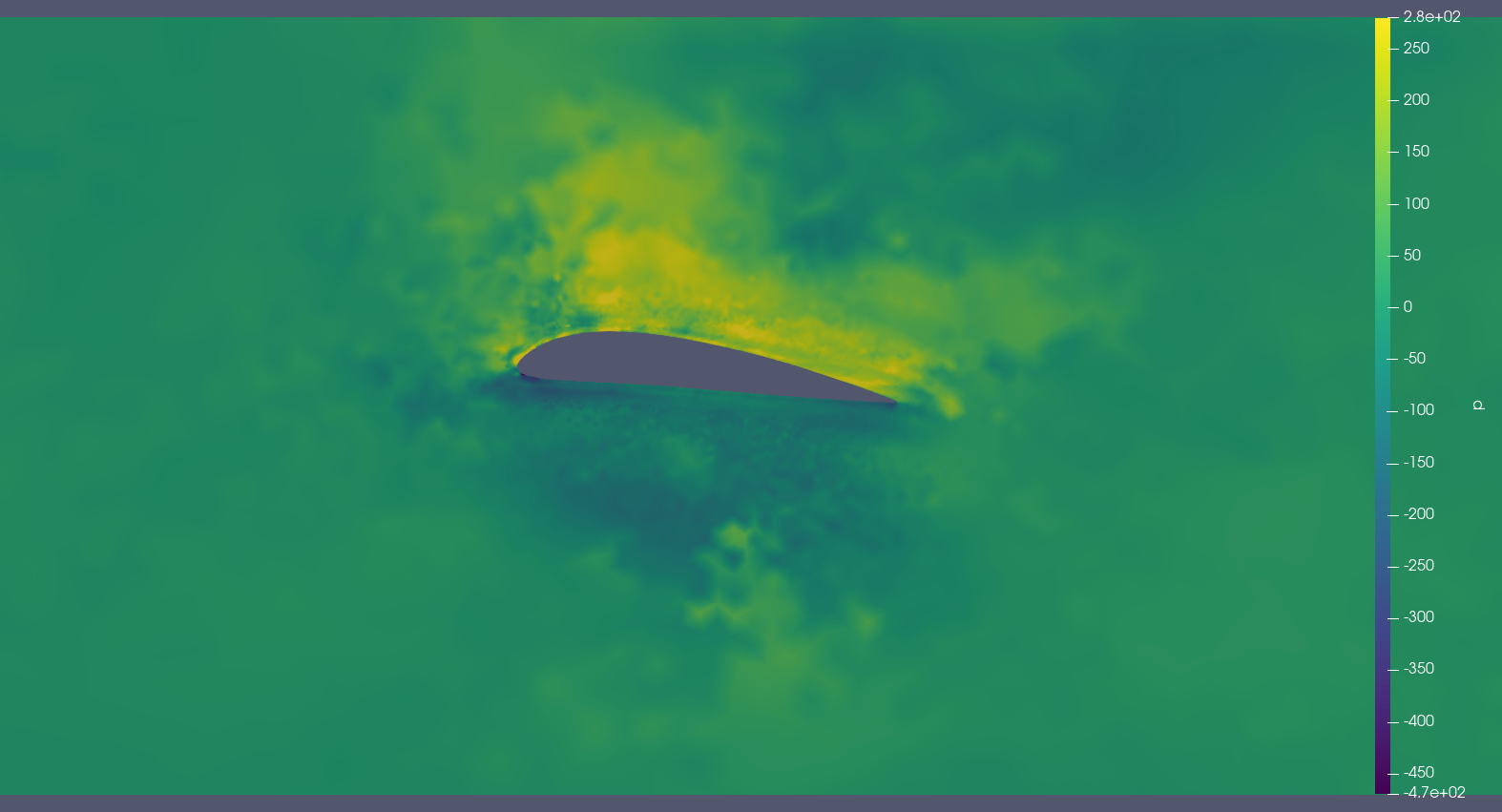}  
            \caption{PointNet++}
    \end{subfigure}
    \begin{subfigure}{0.95\textwidth}
      \centering
      \includegraphics[width=0.5\textwidth]{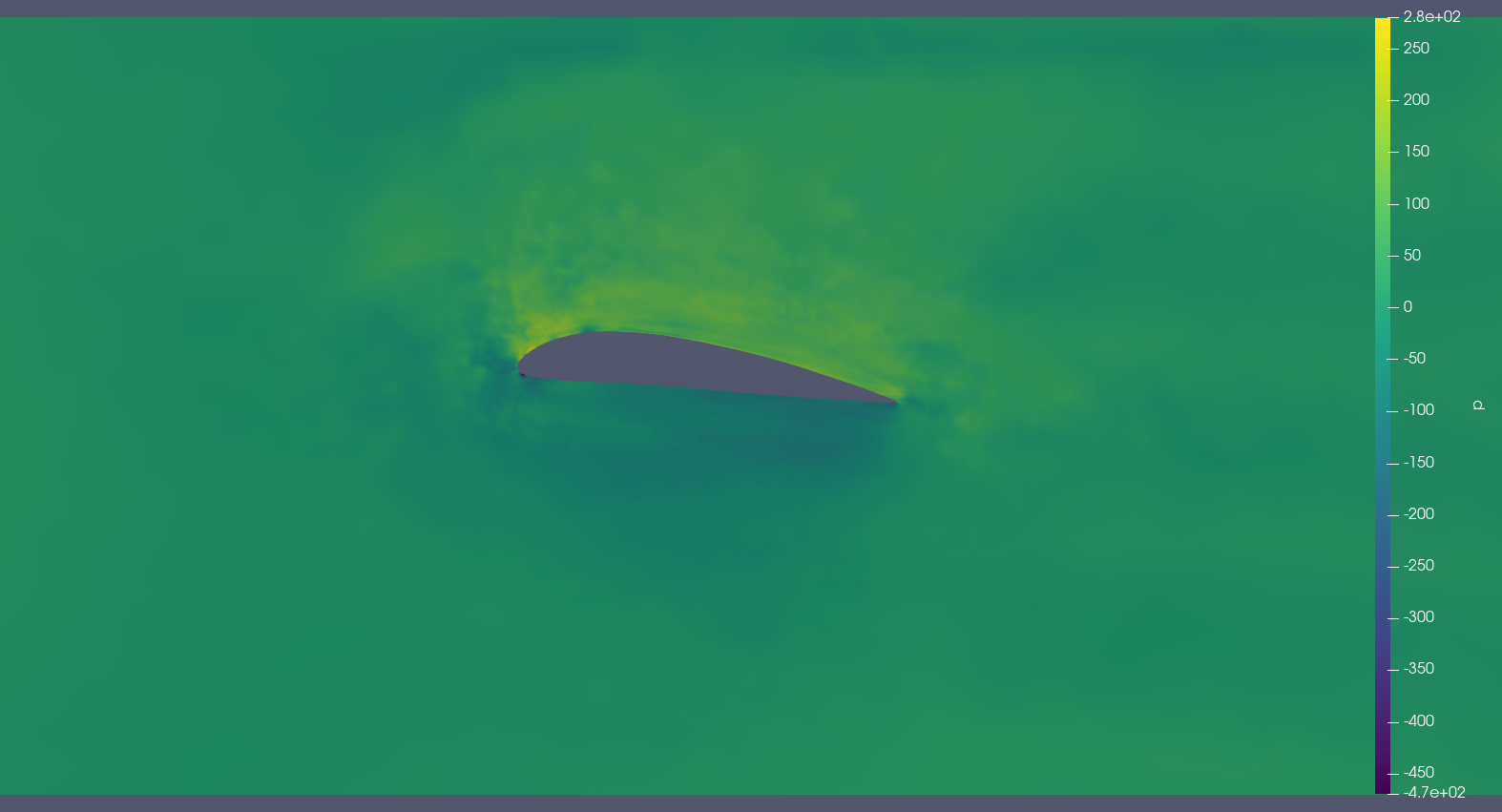}  
            \caption{MGNO}
    \end{subfigure}
    
    \caption{Difference of the pressure field between the ground truth and the different models.}
    \label{fig:qualitative_dp}
\end{figure}

\begin{figure}[ht]
    \centering
    \begin{subfigure}{0.45\textwidth}
      \centering
      \includegraphics[width=\textwidth]{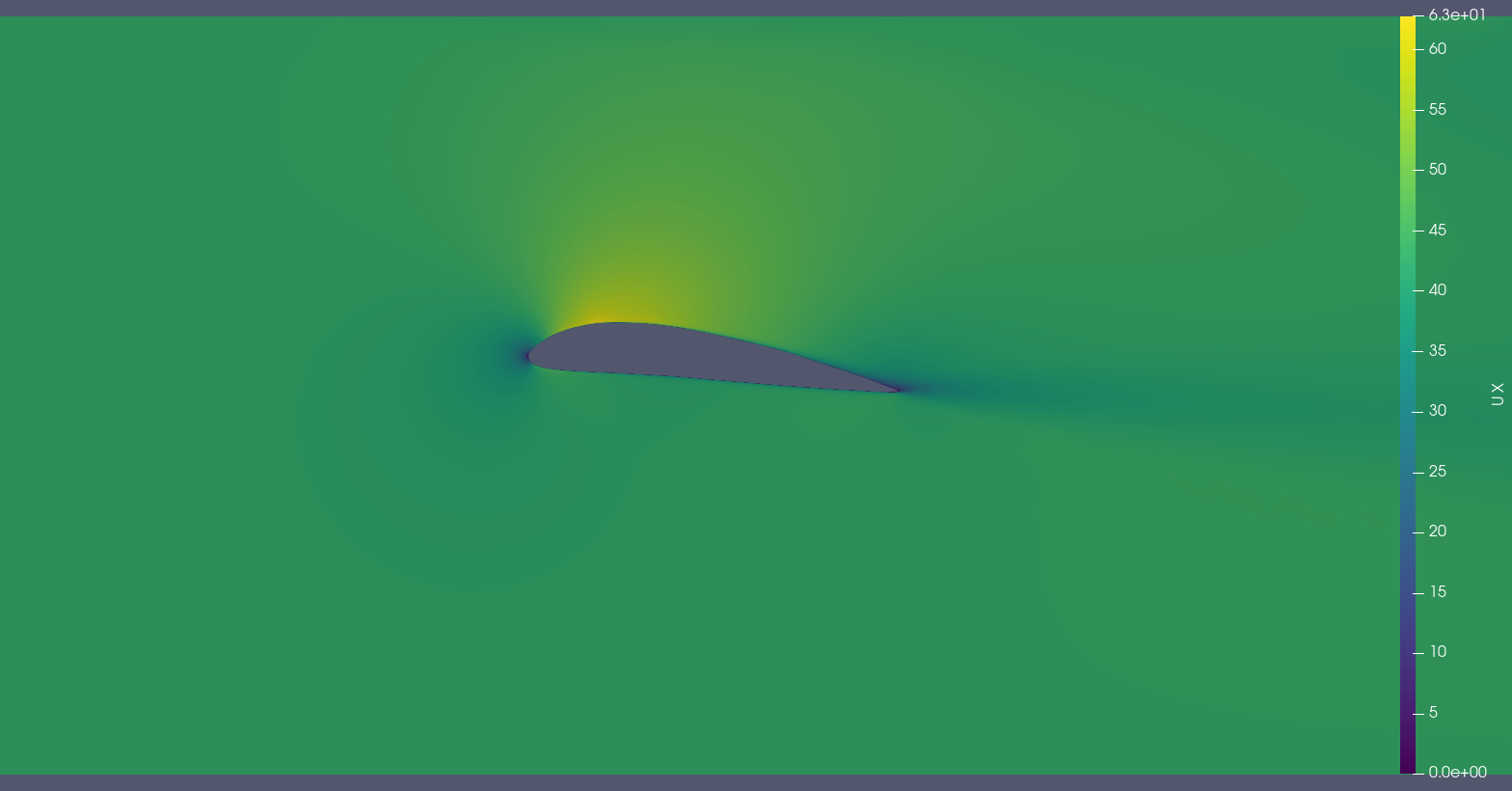}  
            \caption{Ground truth}
    \end{subfigure}
    \begin{subfigure}{0.45\textwidth}
      \centering
      \includegraphics[width=\textwidth]{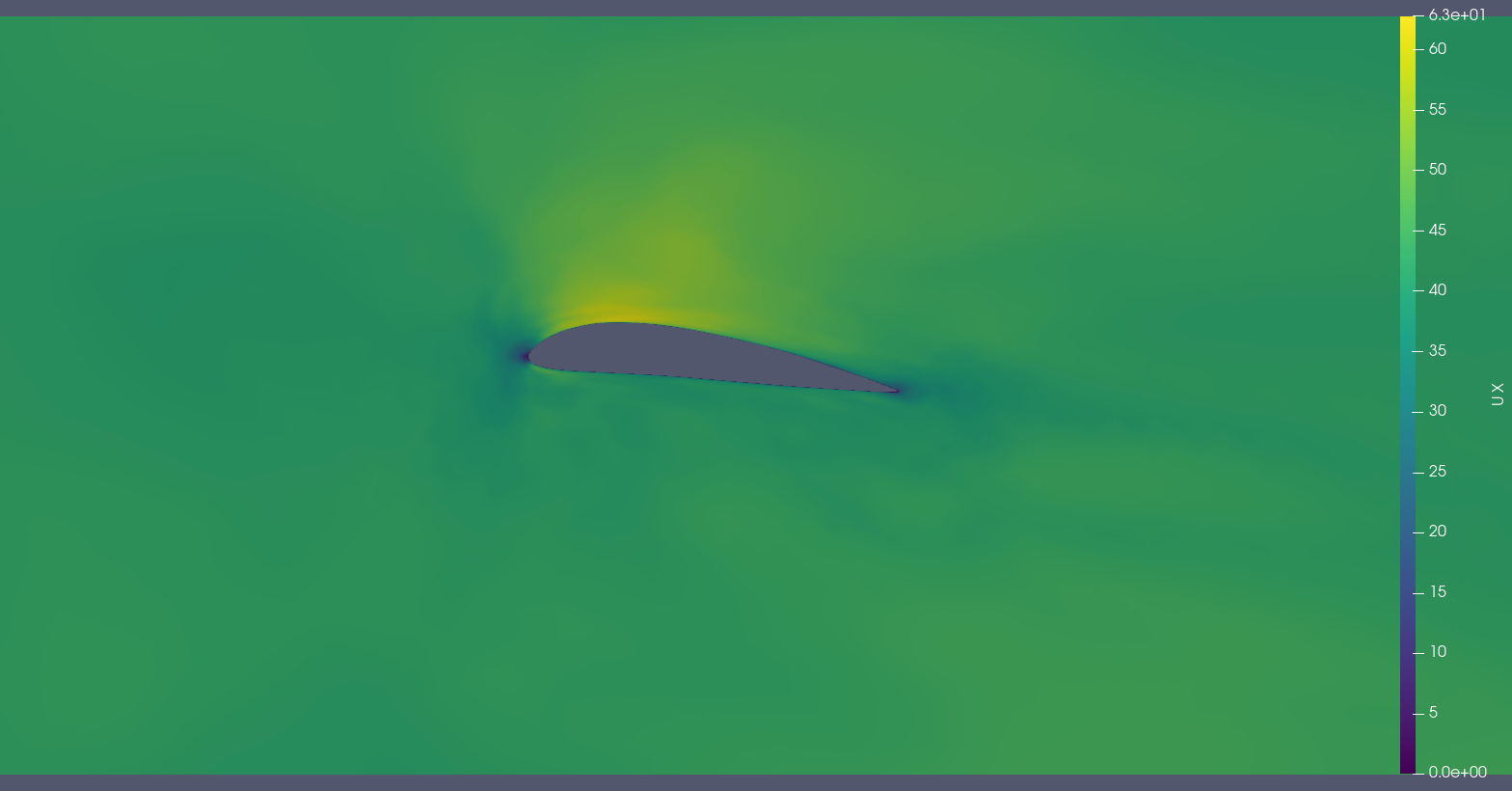}  
            \caption{GraphSAGE}
    \end{subfigure}
    \begin{subfigure}{0.45\textwidth}
      \centering
      \includegraphics[width=\textwidth]{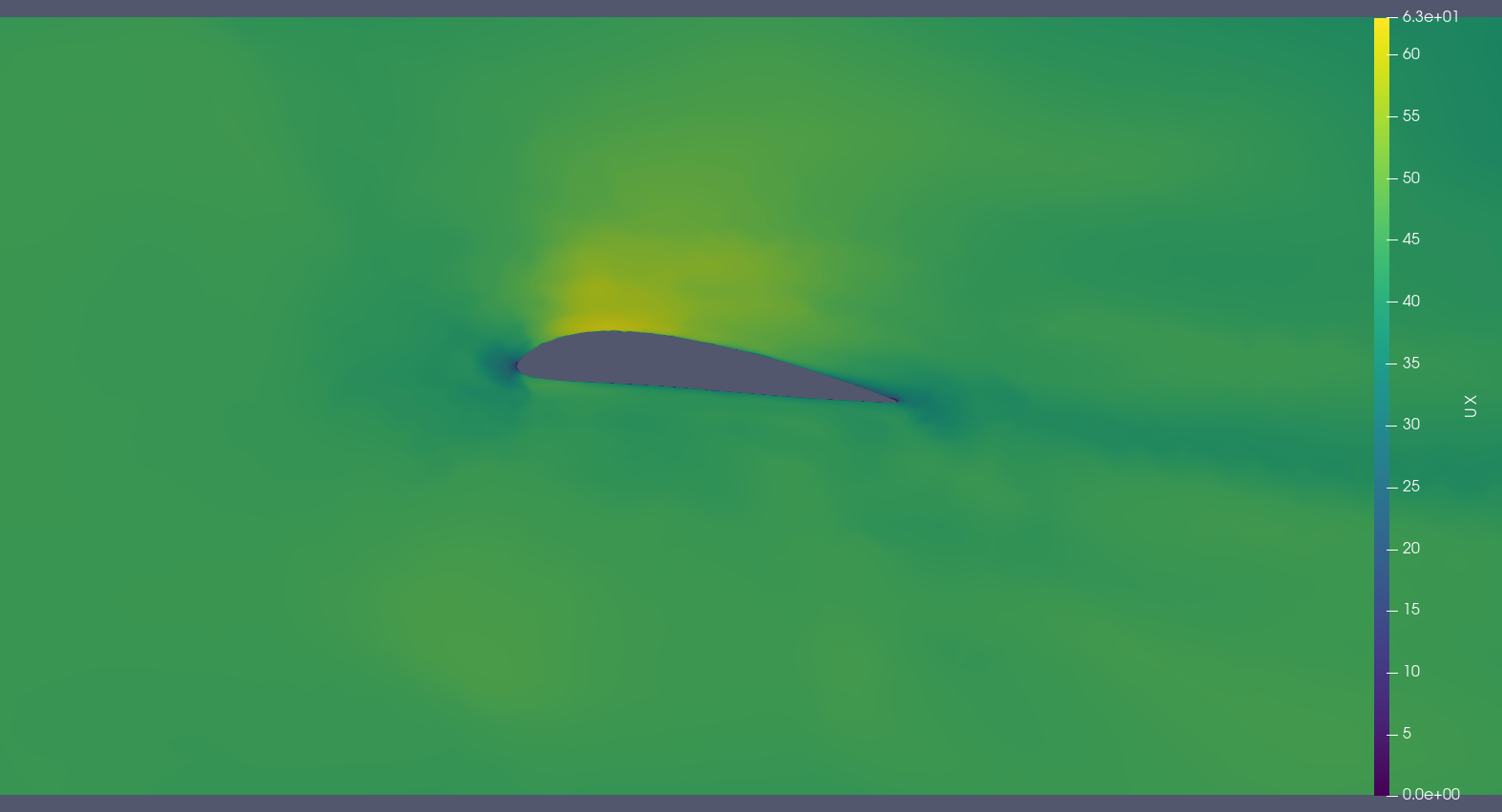}  
            \caption{GAT}
    \end{subfigure}
    \begin{subfigure}{0.45\textwidth}
      \centering
      \includegraphics[width=\textwidth]{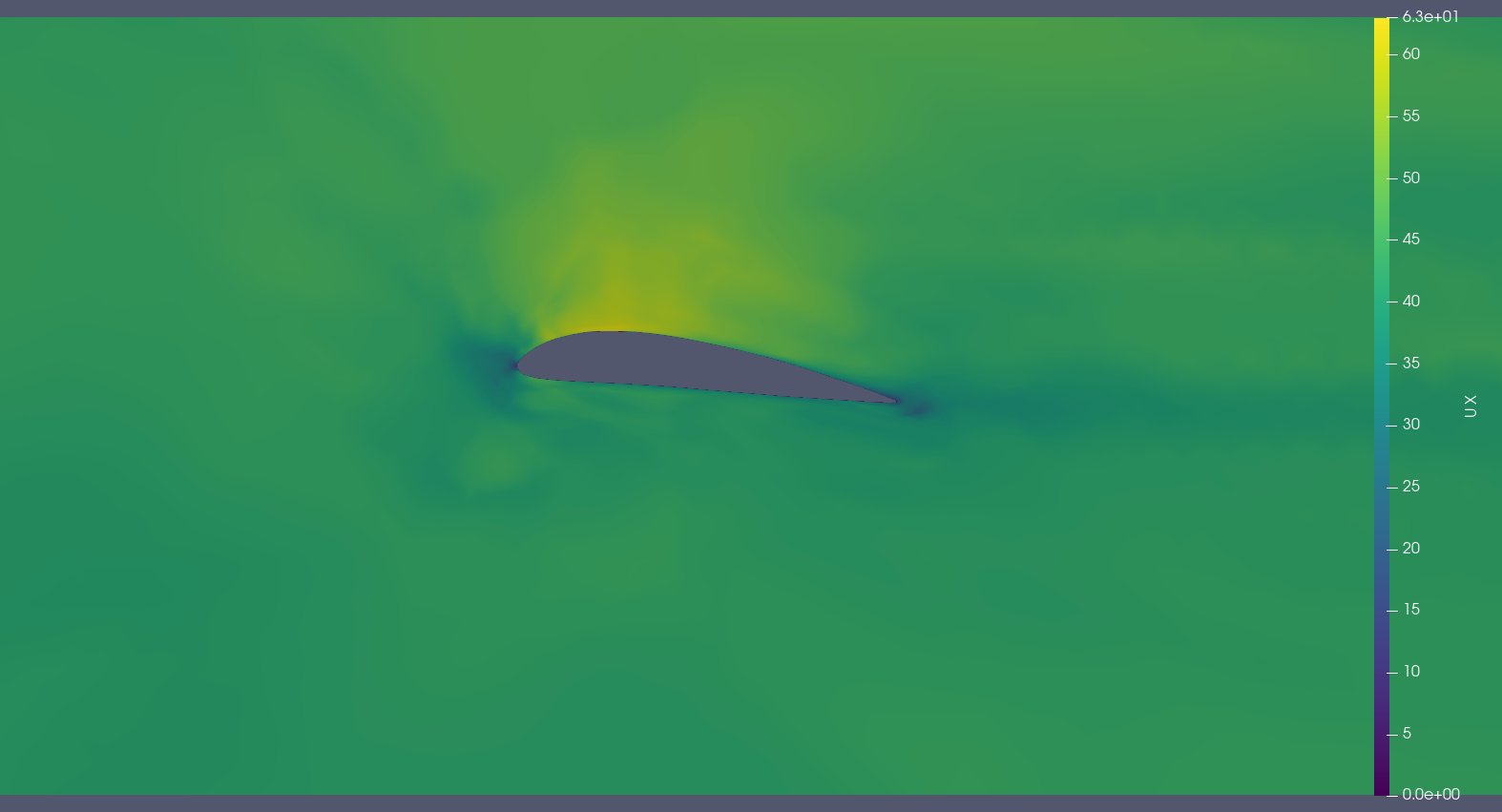}  
            \caption{PointNet}
    \end{subfigure}
    \begin{subfigure}{0.45\textwidth}
      \centering
      \includegraphics[width=\textwidth]{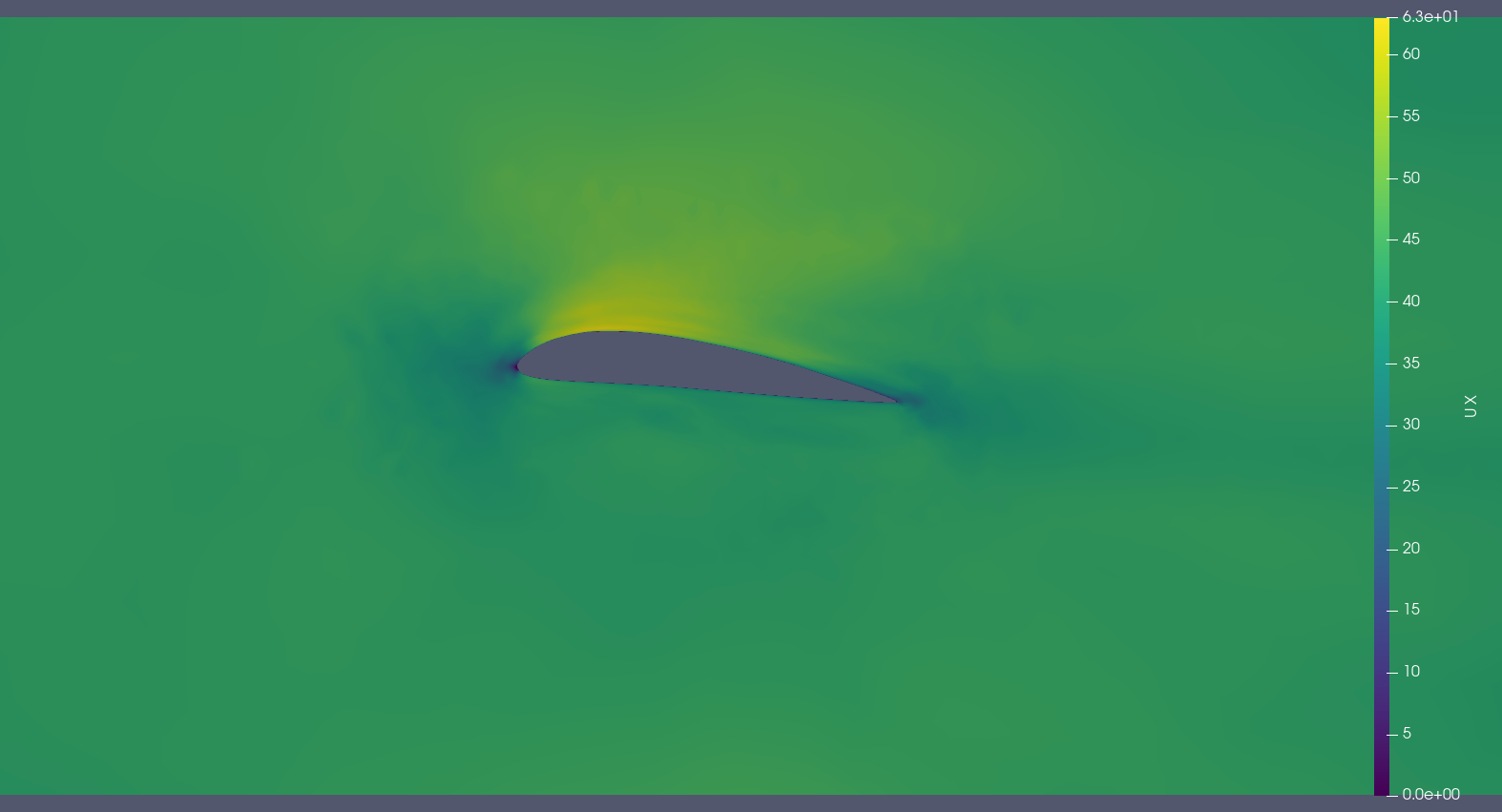}  
            \caption{GNO}
    \end{subfigure}
    \begin{subfigure}{0.45\textwidth}
      \centering
      \includegraphics[width=\textwidth]{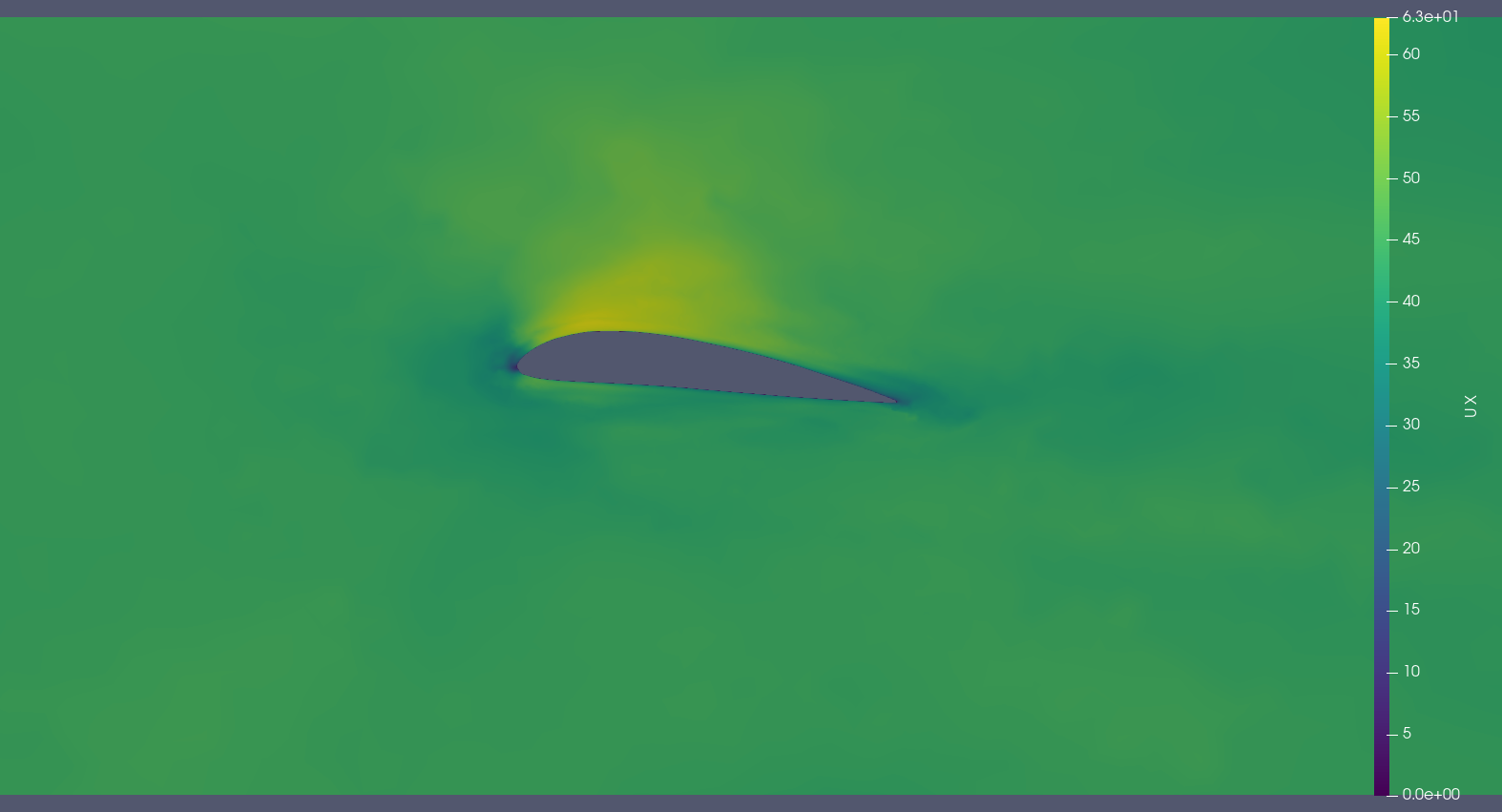}  
            \caption{Graph U-Net}
    \end{subfigure}
    \begin{subfigure}{0.45\textwidth}
      \centering
      \includegraphics[width=\textwidth]{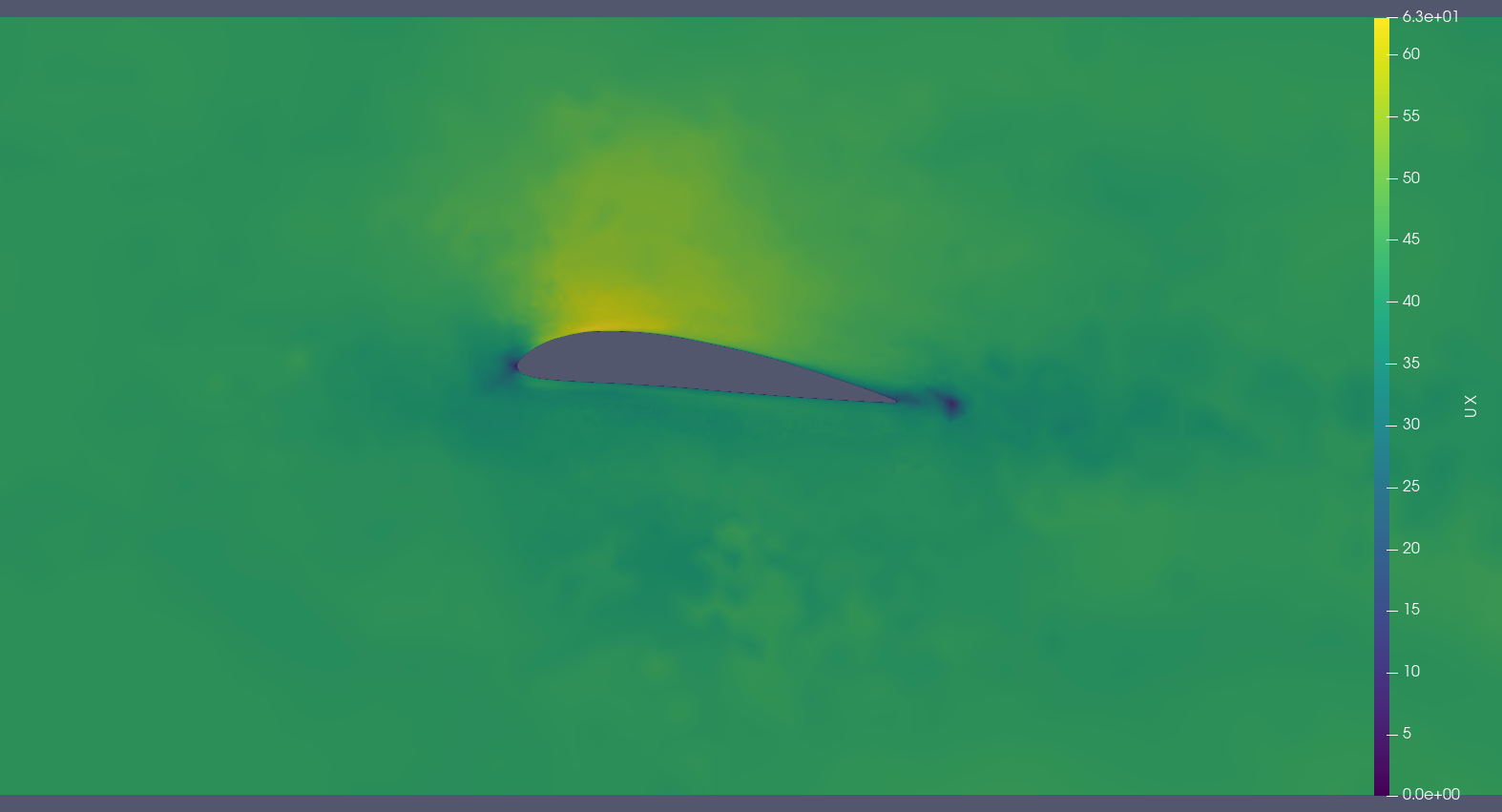}  
            \caption{PointNet++}
    \end{subfigure}
    \begin{subfigure}{0.45\textwidth}
      \centering
      \includegraphics[width=\textwidth]{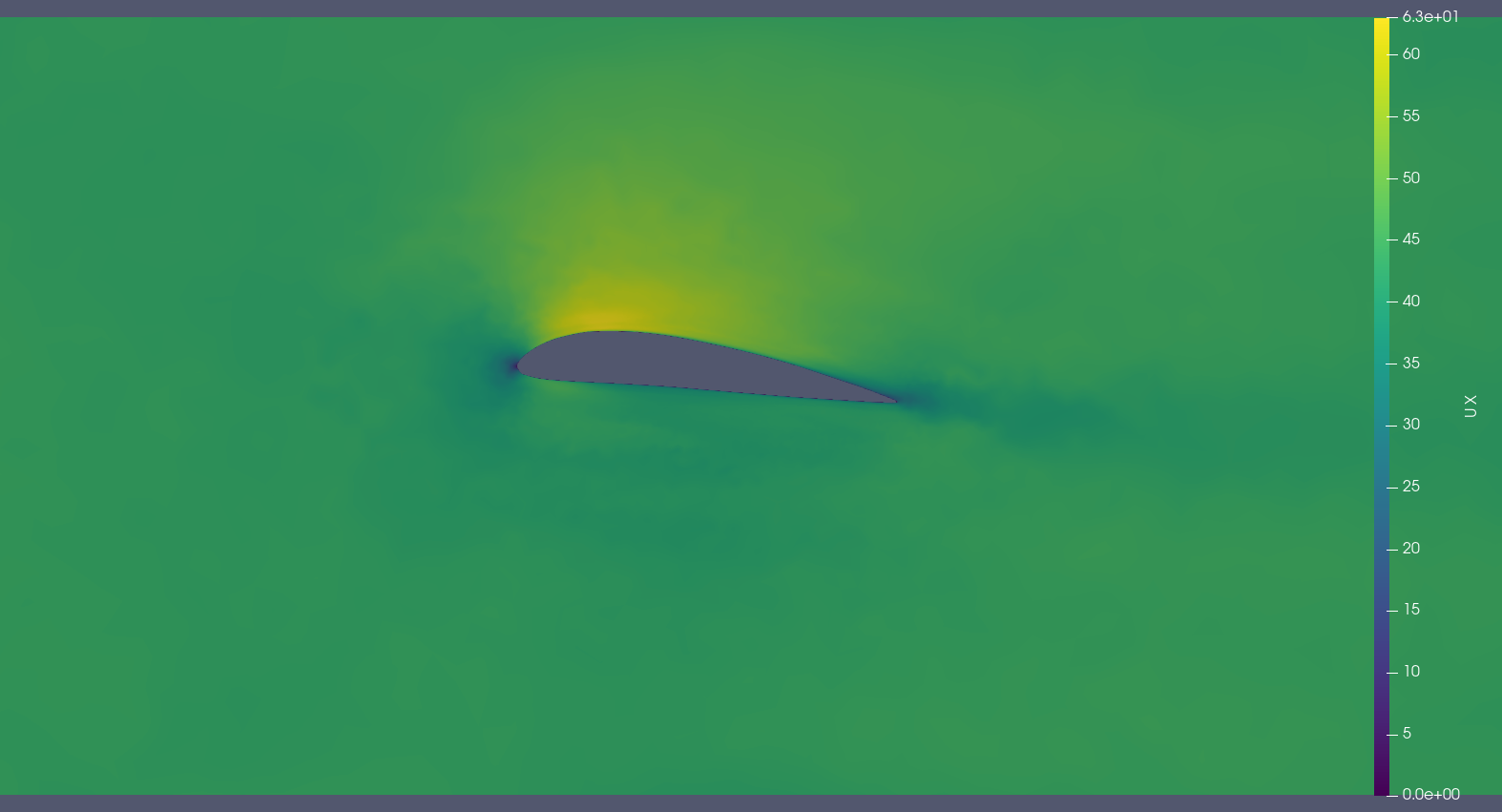}  
            \caption{MGNO}
    \end{subfigure}
    
    \caption{Comparison of the $x$-component of the velocity field for the different models.}
    \label{fig:qualitative_vx}
\end{figure}

\begin{figure}[ht]
    \centering
    \begin{subfigure}{0.45\textwidth}
      \centering
      \includegraphics[width=\textwidth]{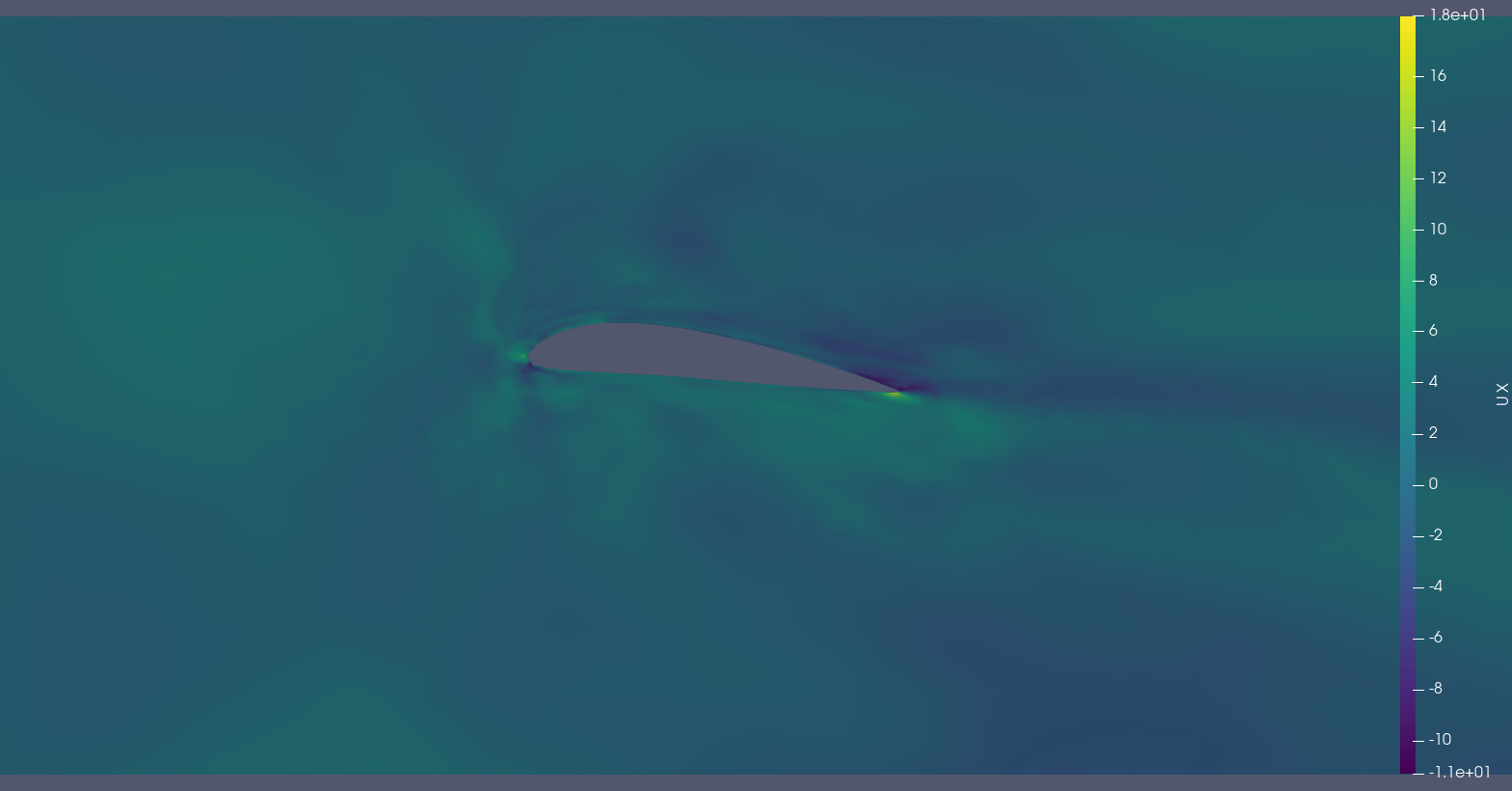}  
            \caption{GraphSAGE}
    \end{subfigure}
    \begin{subfigure}{0.45\textwidth}
      \centering
      \includegraphics[width=\textwidth]{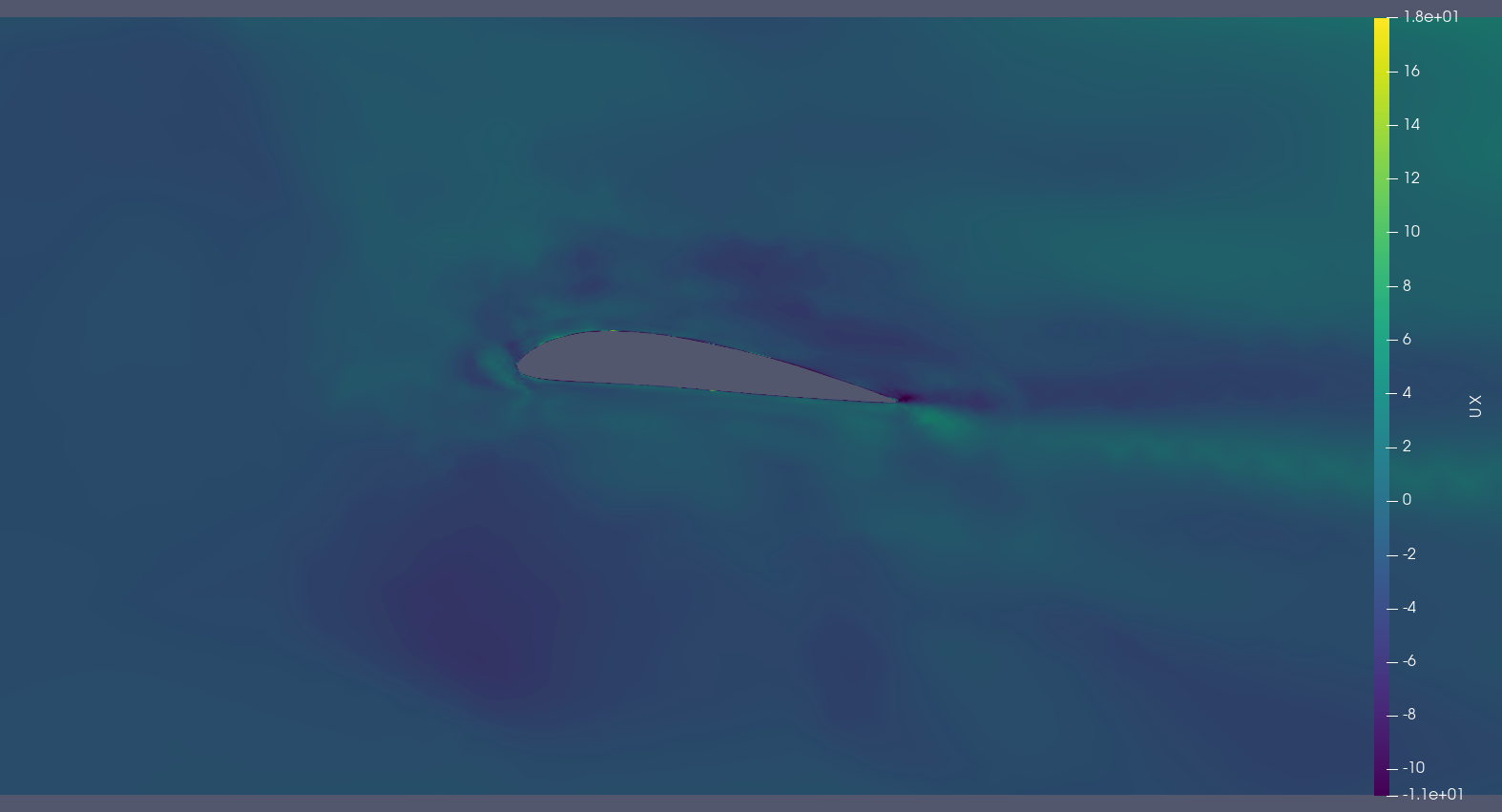}  
            \caption{GAT}
    \end{subfigure}
    \begin{subfigure}{0.45\textwidth}
      \centering
      \includegraphics[width=\textwidth]{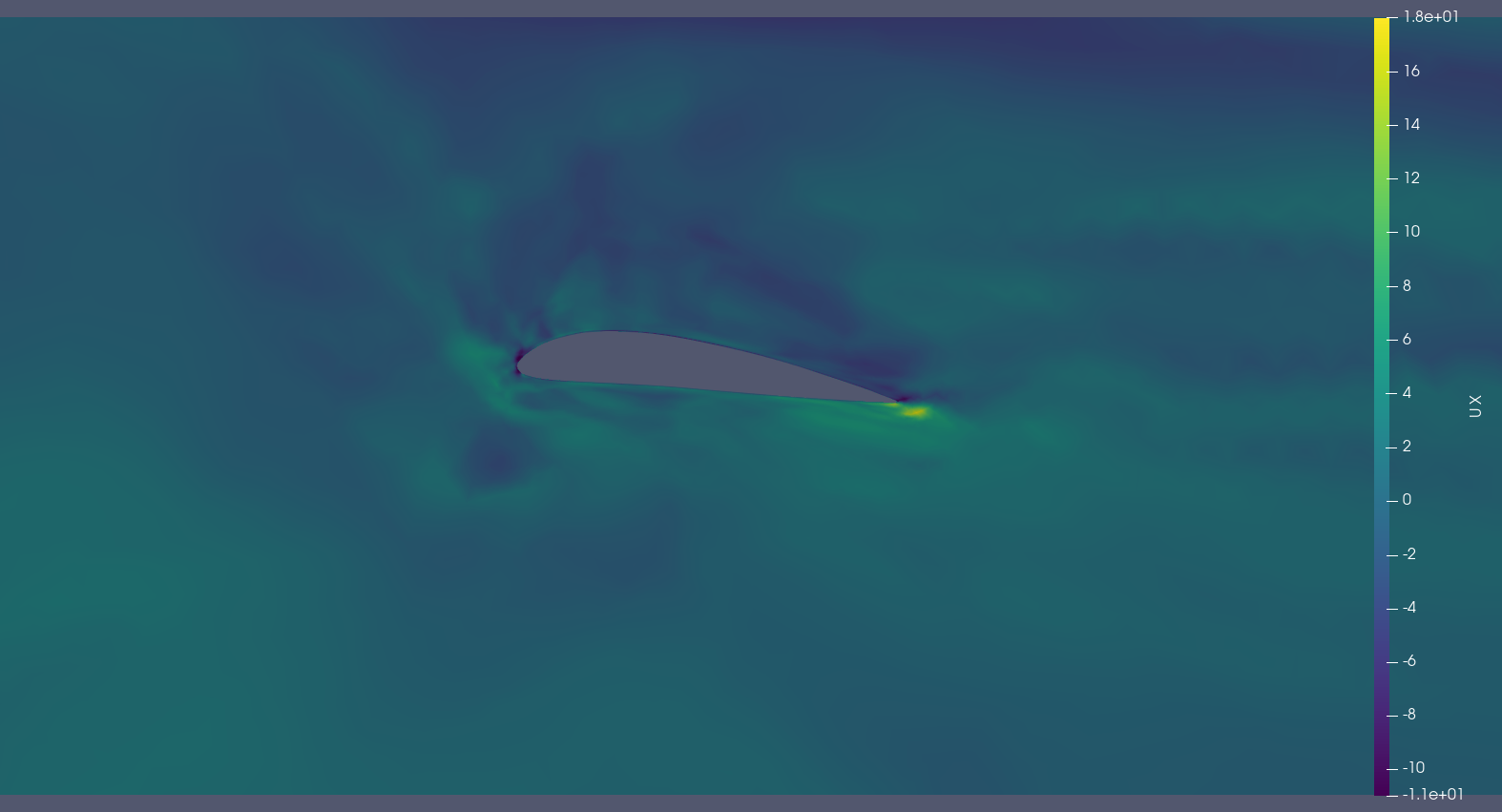}  
            \caption{PointNet}
    \end{subfigure}
    \begin{subfigure}{0.45\textwidth}
      \centering
      \includegraphics[width=\textwidth]{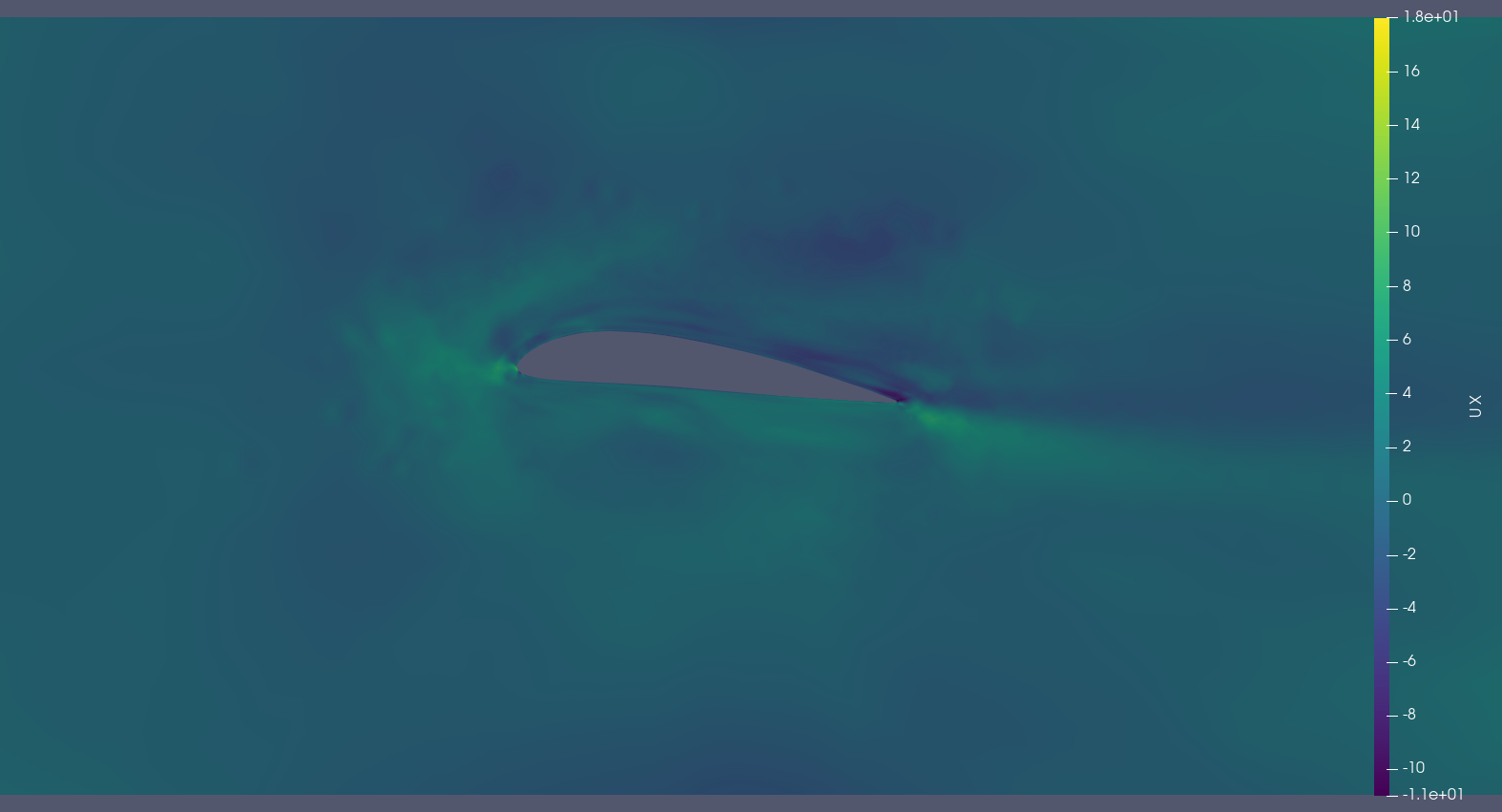}  
            \caption{GNO}
    \end{subfigure}
    \begin{subfigure}{0.45\textwidth}
      \centering
      \includegraphics[width=\textwidth]{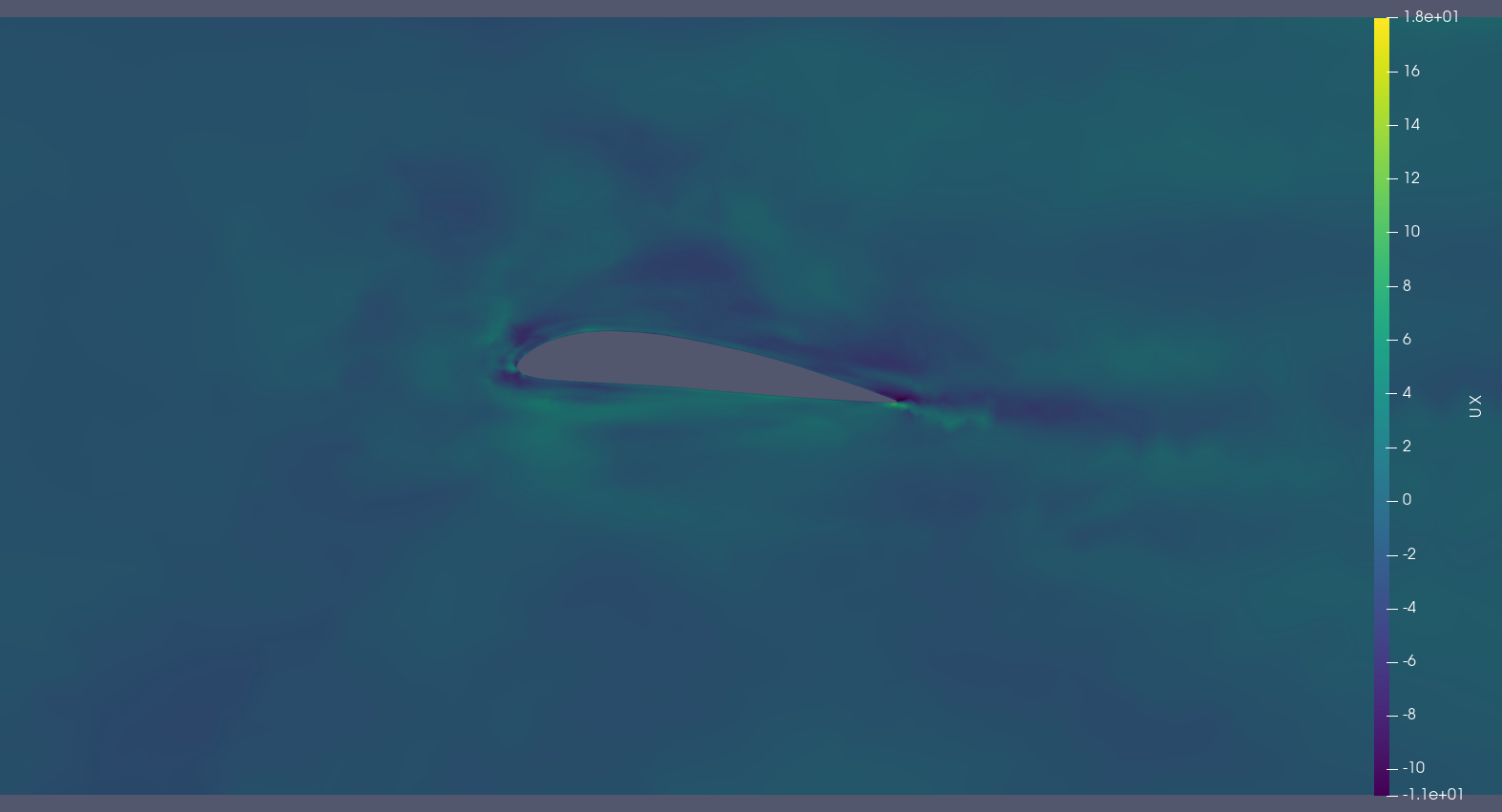}  
            \caption{Graph U-Net}
    \end{subfigure}
    \begin{subfigure}{0.45\textwidth}
      \centering
      \includegraphics[width=\textwidth]{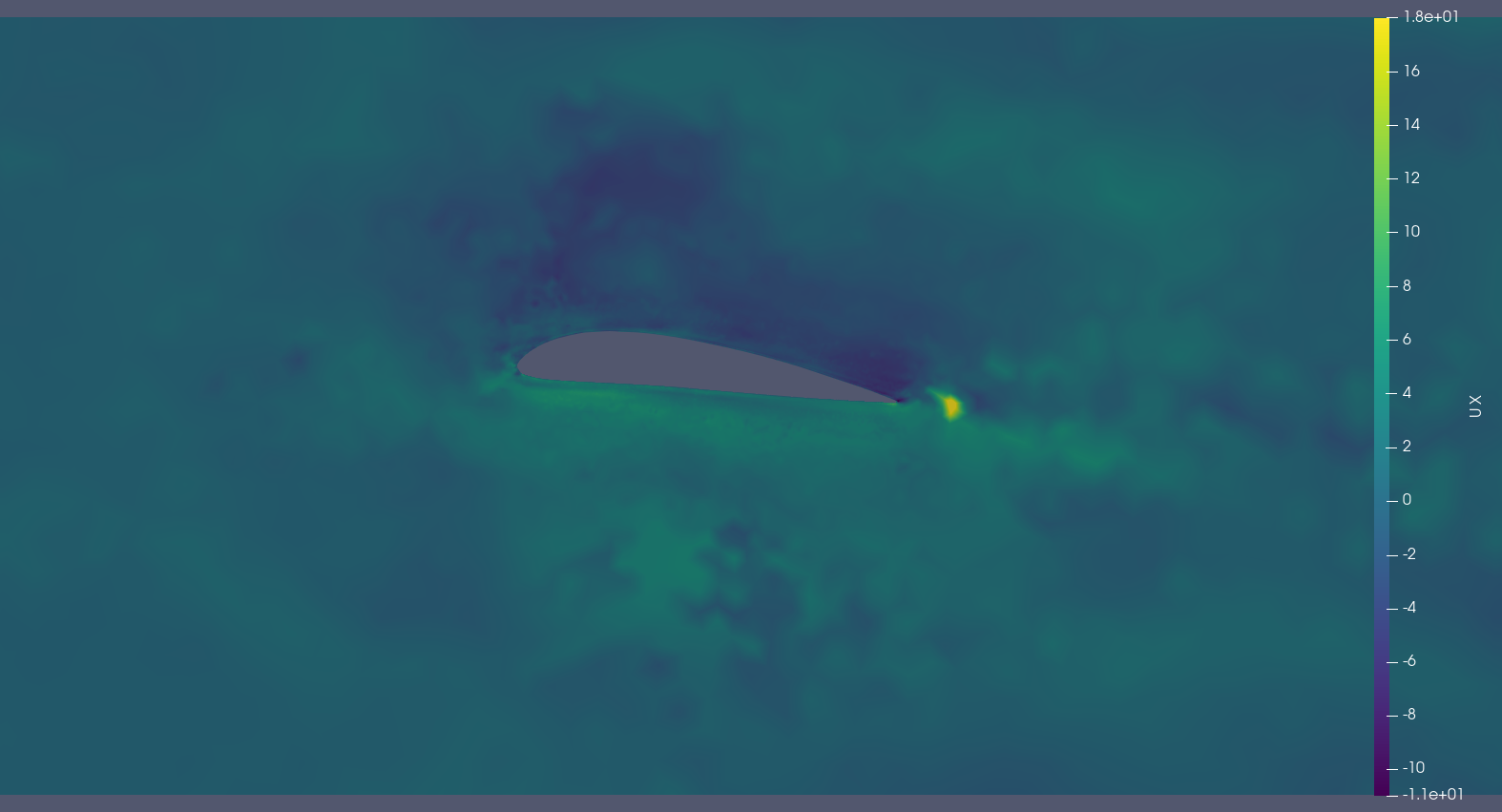}  
            \caption{PointNet++}
    \end{subfigure}
    \begin{subfigure}{0.95\textwidth}
      \centering
      \includegraphics[width=0.5\textwidth]{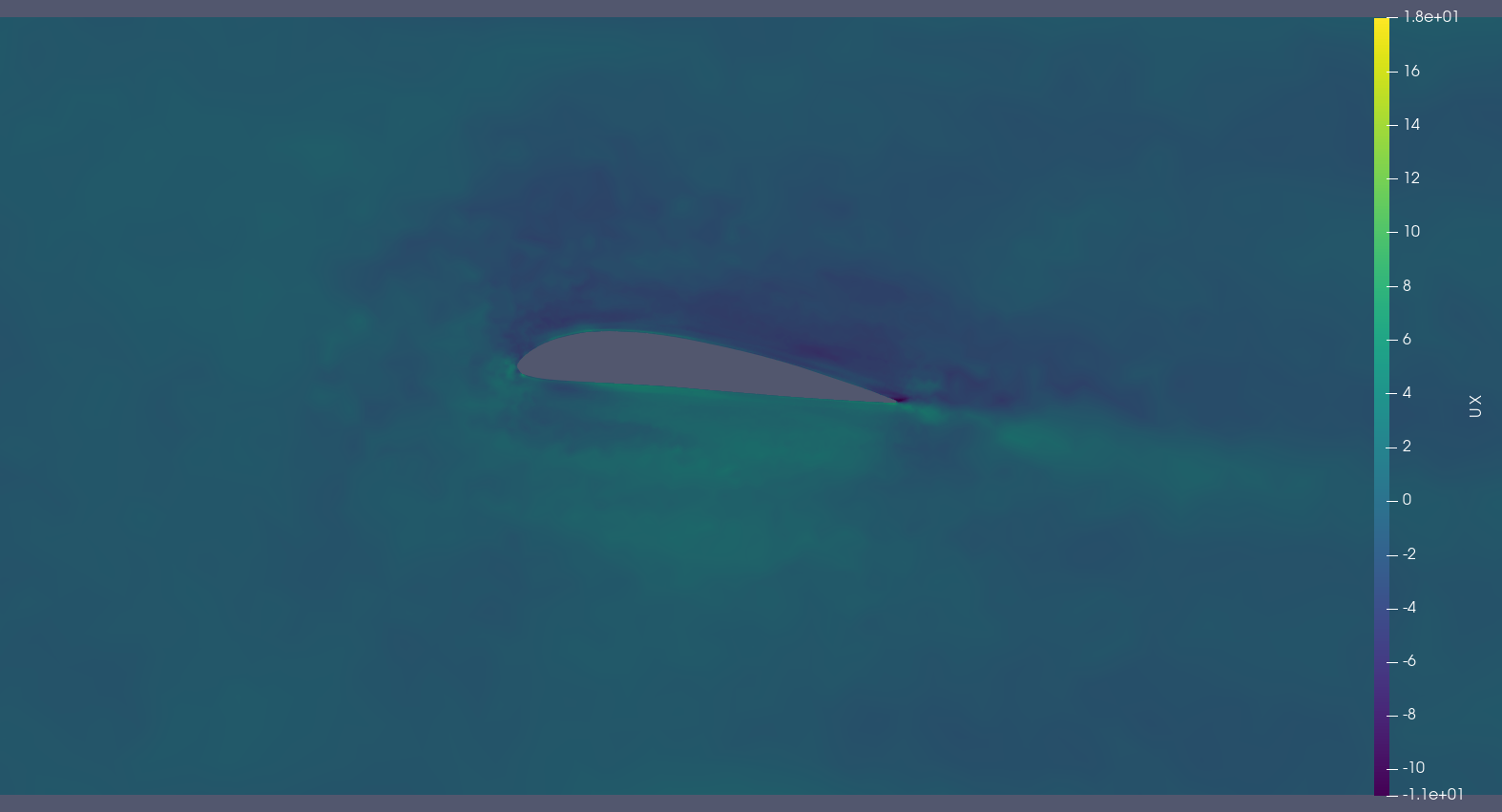}  
            \caption{MGNO}
    \end{subfigure}
    
    \caption{Difference of the $x$-component of the velocity field between the ground truth and the different models.}
    \label{fig:qualitative_dvx}
\end{figure}

In Figure \ref{fig:qualitative_p} and Figure \ref{fig:qualitative_vx}, we show respectively the pressure and the $x$-component of the velocity fields, inferred by the different models compared to the ground truth and in Figure \ref{fig:qualitative_dp} and Figure \ref{fig:qualitative_dvx} we show the difference between the ground truth and the predicted fields.

\subsection{Single-scale models}
\subsubsection{GraphSAGE}\label{sec:graphsage_cfd_mesh_radius_graph}
\begin{figure}
    \centering
    \includegraphics[width=\linewidth]{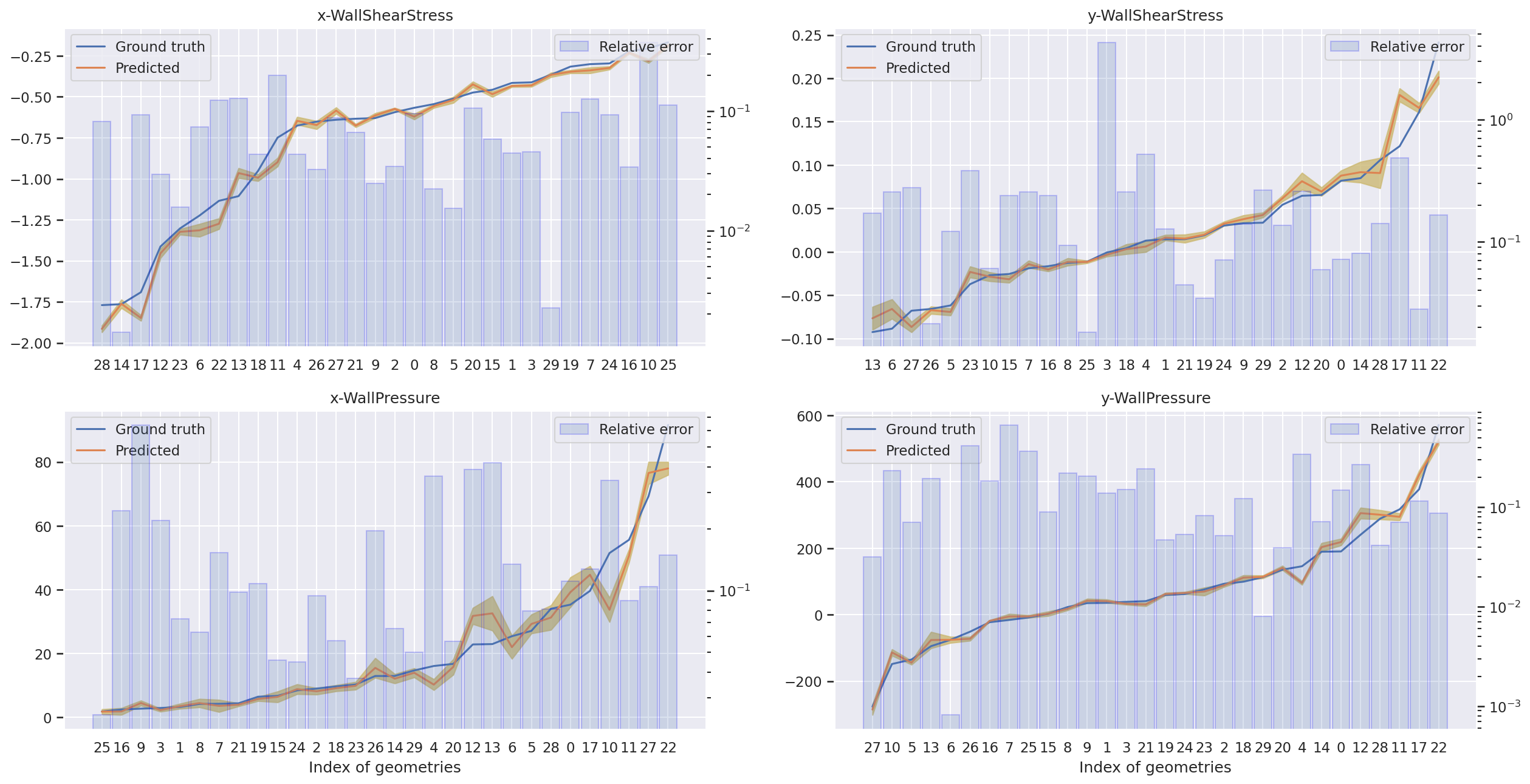}
    \caption{Order plot over the different samples in the test set of the stress forces for the GraphSAGE model. The relative errors are given in logarithmic scale with respect to the mean value of the stress forces.}
    \label{fig:SAGE_global}
\end{figure}

The GraphSAGE layer \citep{SAGE} is the basic inductive type layer, it gives us a first indicator of the difficulty of the task. We used a 4-layers GraphSAGE network $8-64-64-64-8$ channels, each layer is followed by a batchnorm layer to make the optimization problem easier and a ReLU activation function. Also, as the number of nodes is still pretty small in our dataset, the CFD mesh holds in the memory of our GPU. Hence, we compared the scores of the same model trained over the CFD mesh and with our downsampling technique using the radius graph. Table \ref{ap:GSAGE_score} reports the results of this comparison. We observe that even though the CFD mesh includes 5 to 10 times more points than the radius graph during training, the model has similar scores with the radius graph setting. In Figure \ref{fig:SAGE_global}, we also show the inferred WSS and WP with the radius graph methods and compare it with the ground truth.

\begin{table}
    \caption{Scores of the GraphSAGE model on the radius graph and the CFD mesh.}
    \resizebox{0.99\textwidth}{!}{
    \centering
    \begin{tabular}{c|c|c|c|c|c|c}

        \bf Graph & \multicolumn{2}{c|}{\bf Local metrics} & \multicolumn{4}{c}{\bf Integral geometry forces} \\
        & $\mathcal{L}_\mathcal{V}$ & $\mathcal{L}_\mathcal{S}$ & x-WSS & y-WSS & x-WP & y-WP \\
                & $\left(\times 10^{-2}\right)$ & $\left(\times 10^{-2}\right)$ & $\left(\times 10^{-3}\right)$ & $\left(\times 10^{-4}\right)$ & $\left(\times 10\right)$ & $\left(\times 10^{2}\right)$ \\
        \hline
        &  &  &  &  &  & \\
        Radius graph & \textbf{2.94 $\pm$ 0.20} & 4.71 $\pm$ 0.35 & 5.12 $\pm$ 0.65 & \textbf{2.93 $\pm$ 0.50} & \textbf{3.47 $\pm$ 0.89} & \textbf{6.10 $\pm$ 1.13} \\
        CFD mesh & 3.00 $\pm$ 0.24 & \textbf{4.54 $\pm$ 0.40} & \textbf{3.47 $\pm$ 1.05} & 3.41 $\pm$ 0.71 & 6.29 $\pm$ 1.12 & 6.94 $\pm$ 0.94
    \end{tabular}
    }
    \label{ap:GSAGE_score}
\end{table}

\subsubsection{GAT}
The GAT layer \citep{GAT} is designed for inductive tasks too but is often better performing than GraphSAGE. We used a 4-layers GAT network $8-64-64-64-8$ channels, each layer is followed by a batchnorm layer and a ReLU activation function.

\subsubsection{PointNet}
PointNet \citep{pointnet} is not a GNN but a model that operate on point clouds. It is a two scales architecture that is better equipped to handle long-range interactions between nodes compared to the aforementioned GNNs. Moreover, it can be built with a GraphSAGE or a GAT layer for the representation task. We chose to keep a MLP for the representation task as it gives similar scores and is faster to train. For the architecture, see \citep{pointnet} for the segmentation task, we just changed the output in order to fit with our problem and get rid of the batchnorm layers and dropout as it was performing poorly with.

\subsection{Multi-scale models}
The resolution of PDEs may involve long-range interactions where all the points of the domain influence the solution at a certain position. It is difficult to take into account those interactions between far-away nodes in architectures such as the GraphSAGE or GAT as it would imply stacking a consequent number of layers (equivalent to the diameter of a graph) which would make the training process difficult. Multi-scale models can overcome this issue by representing the signal at different scales allowing long-range interactions by coarsening the input graphs or point clouds.

\subsubsection{Graph-Unet}
The Graph-Unet \citep{GraphUNet} is the archetypal multi-scale architecture. It extends the well-known U-Net architecture \citep{Ronneberger2015UNetCN} initially designed  for image segmentation, to graph data. Our Graph-Unet makes use of GraphSAGE layers instead of GCN \citep{Kipf:2016tc} layers as in the historical paper. Moreover, we did not use the gPool technique developed in the Graph-Unet as we found that it is not robust to the downsampling. We replaced it by a simple random downsampling and nearest neighbour upsampling. Our architecture is composed of 5 scales with downsampling ratio of $3/4-3/4-2/3-2/3$ and at each scale a radius graph is built with radius of $0.1-0.2-0.5-1-10$ where the last radius is taken big enough for the graph to be totally connected. In the training process, each radius graph is set to produce a maximum of 64 neighbors whereas for inference we raise this number to 512.

The architecture is composed of 9 layers, each followed by a batchnorm layer, a ReLU activation function and a downsampling or upsampling layer (for all but the last block). The number of channels is doubled at each scale starting at 8 and the intra-scale information of the Graph U-net are aggregated via concatenations. Hence, the number of channels is $8-16-32-64-128$ in the downward pass and $192-96-48-24-8$ in the upward pass.

\subsubsection{PointNet++}
The PointNet++ \citep{pointnet++} is a multi-scale extension of the PointNet model discussed above. It allows refinement in the global representation of the PointNet by using multiple PointNet on local areas leading to a multi-scale representation of the task. Our architecture is the same as described in the original paper for the segmentation task. We chose 7 scales with downsampling ratio of $3/4-3/4-2/3-2/3-3/4-2/3$ and on each scale a radius graph is built with radius of $0.1-0.2-0.4-0.8-1.6-10$ where the last radius is taken big enough for the graph to be totally connected. In the training process, each radius graph is set to produce a maximum of 64 neighbors whereas for inference we raise this number to 512. No dropout is used.

\subsection{Graph Neural Operators}
Neural operators are a new paradigm in DL that aim at approximating operators between Hilbert spaces instead of applications between real and finite dimensional vector spaces \citep{kovachki2021universal}. Those models are often composed of an encoder which is in charge of finding a finite dimensional representation of the infinite dimensional input space, an approximator that transforms this representation and a decoder that maps this finite representation of the solution to the infinite dimensional output space. 
\newpage
\subsubsection{Graph Neural Operator (GNO)}
\begin{wrapfigure}{r}{0.4\textwidth}
    \centering
    \includegraphics[height = 0.5\textheight]{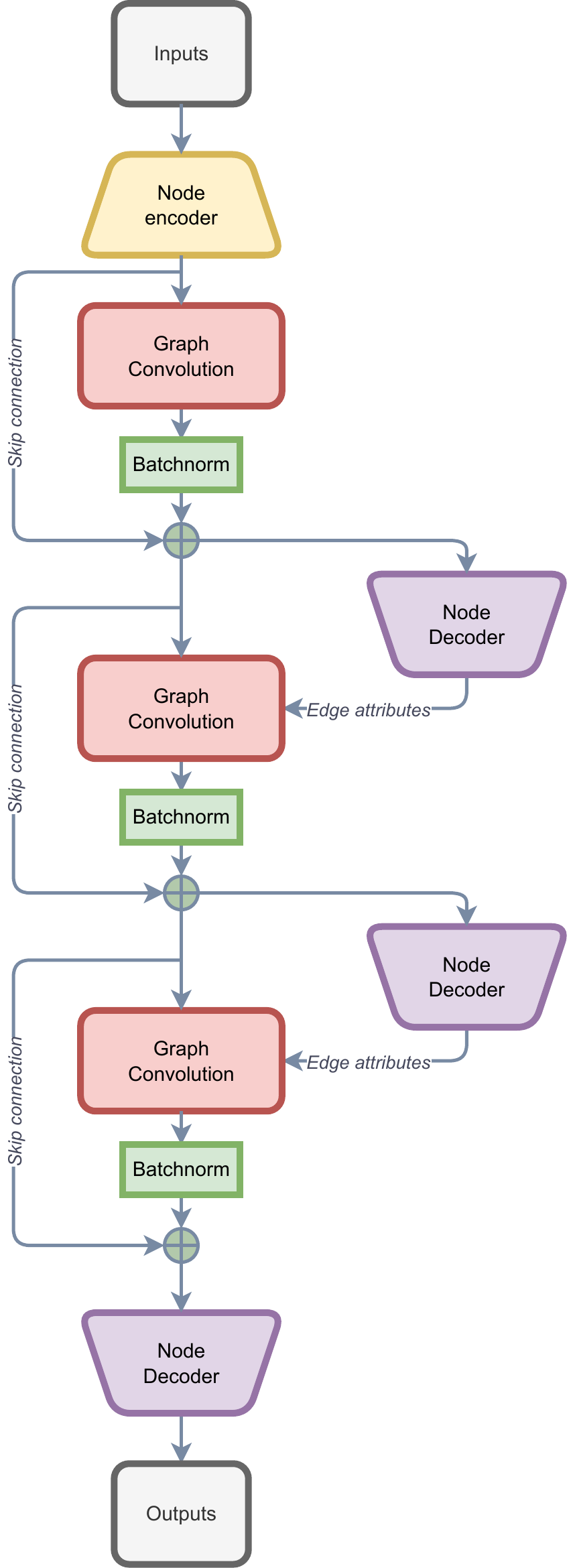}
    \caption{Representation of the GNO architecture for $T = 3$.}
    \vspace{-1cm}
    \label{fig:GKN_archi}
\end{wrapfigure}

Our model is a modulation of the architecture proposed in \citep{GKN}. It is a recurrent network that mimics the resolution of Poisson's equation through a convolutional operator and repeats it several times to approximate the solutions. We can write it as:
\begin{align}
    h^t_i = \frac{1}{|\mathcal{N}_i|}\sum_{j\in\mathcal{N}_i} \kappa_{\theta_k}(e^{t-1}_{ij})h^{t-1}_j + h^{t-1}_i
\end{align}
where $\mathcal{N}_i$ is the set of neighbors of the node $i$, $\kappa_{\theta_k}$ a neural network, $e^{t}_{ij}$ the attributes of the edge between the nodes $i$ and $j$ at iteration $t$ and $h^t_i$ the hidden state at iteration $t$ and node $i$. We also use an encoder and a decoder at the beginning and the end of our network, such that $h^0_i = \phi_{\theta_e}(x_i)$ and $\hat{u}(x_i) = \psi_{\theta_d}(h^T_i)$ for a network with $T$ iterations, where $\phi_{\theta_e}$ and $\psi_{\theta_d}$ are also both neural networks. Note that just after each graph convolution block, we use a batchnorm layer. The edge attributes $e^t_{ij}$ are defined as a vector containing the relative position, velocity and pressure between nodes $i$ and $j$ at iteration $t$, the signed distance function of node $i$ and $j$ and the inlet velocity. The computation of the velocity field and pressure field at an intermediate iteration is done through the decoding of the hidden state at that iteration. Also, a multiscale architecture has been designed with the same philosophy in order to capture long-range interactions between nodes without blowing up the numerical complexity. In Figure \ref{fig:GKN_archi} we depicts our modified Graph Neural Operator (GNO) architecture.\\
For the kernel in the graph convolution, we used a 4-layers MLP with $8-64-64-64-64$ neurons and ReLU activation function. This kernel takes as inputs the edge attributes of the graph and outputs a convolution matrix of size $8\times 8$.

\newpage

\subsubsection{Multipole Graph Neural Operator (MGNO)}
\begin{wrapfigure}{r}{0.5\textwidth}
    \centering
    \includegraphics[width = 0.5\textwidth]{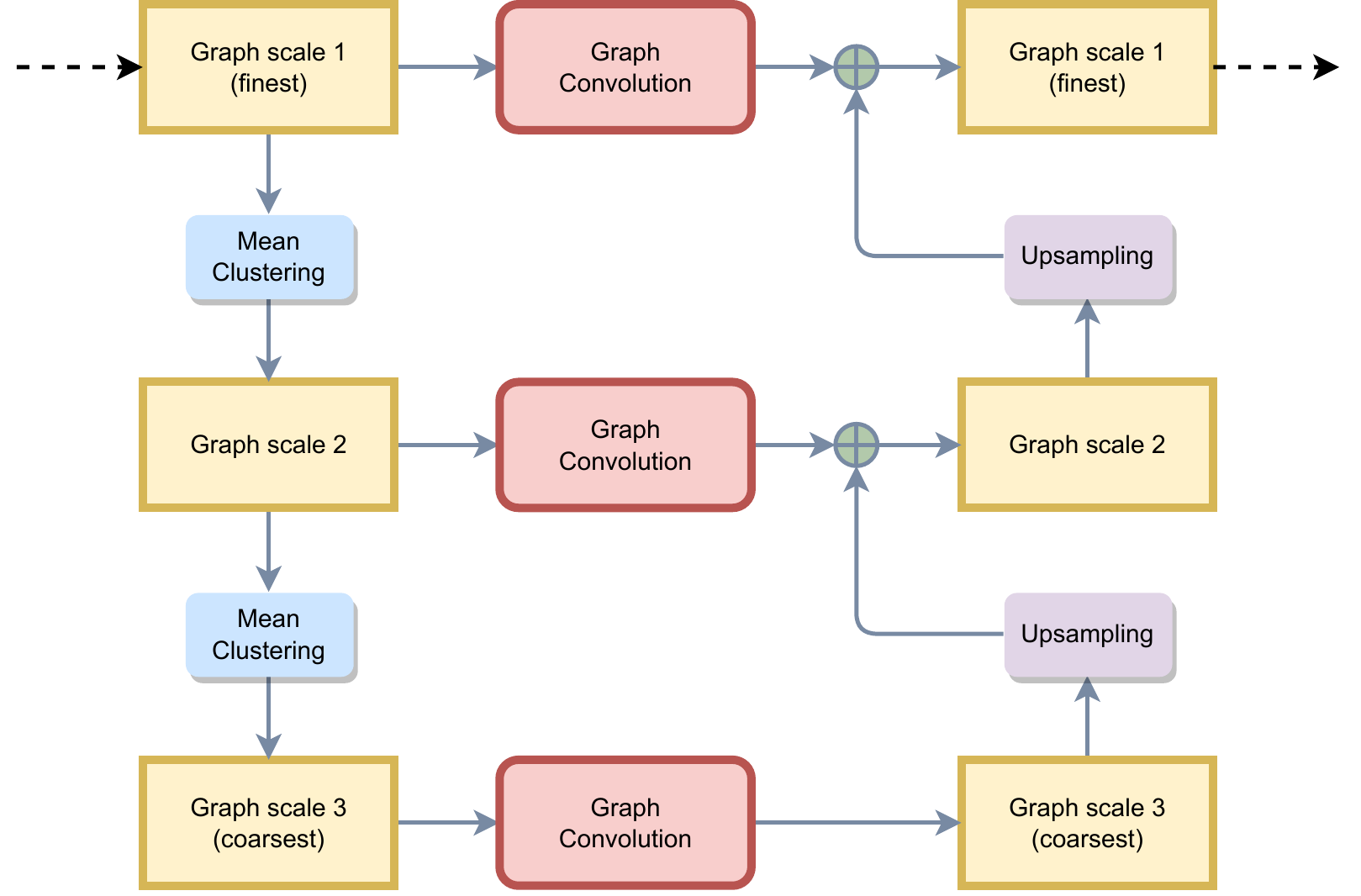}
    \caption{Representation of the MGNO convolution block with 3 scales.}
    \vspace{-0.2cm}
    \label{fig:MGKN_block}
\end{wrapfigure}

We also propose a multiscale version of the GNO, namely the Multipole Graph Neural Operator (MGNO) which is a modulation of \citep{MGKN}. The architecture follows the same form as depicted in Figure \ref{fig:GKN_archi} but the graph convolution here is different from the one in the GNO. We replaced the multipole technique used in the original paper to a more classical U-Net like architecture for the block of convolution. Figure \ref{fig:MGKN_block} depicts the details of the block. A convolution is done at multiple scales by downsampling the input graph via a mean pooling. Then, these convolutions are aggregated scale by scale through a nearest neighbors upsampling. At each scale, a kernel is needed for the convolution. All the kernels are 4-layers MLP with $8-64-64-64-64$ neurons and ReLU activation function as in the GNO model.

\section{Results discussion}\label{sec:result_discussion}
In this section, we discuss the results reported in Table \ref{total_score}. First remark is that multi-scale models do not necessarily perform better than single-scale models. A second remark is that the MSE over the points of the domain is not necessarily a good proxy for good performance on the stress forces. Actually, a simple GraphSAGE model gives similar results compared to more complicated architectures such as GNO or MGNO on stress forces whereas the latter gives better results for the loss function. The training process of the PointNet and PointNet++ networks seems less robust compared to the GraphSAGE or the GNO/MGNO one. However, no model outperforms all the other in every aspect. MGNO yields the best results in all our metrics but the $y$ component of the wall pressure may suffer from scalability problem as it is memory and time consuming to train it. Hence, in this supervised problem settings, we cannot conclude on the efficiency of a particular design such as classical GNNs, neural operators or network acting on point clouds.

\section{Stress forces}\label{sec:drag_lift}
We call \emph{stress} the force $\sigma(n)$ that is acting (by contact) on a surface of unit normal $n$. For an infinitesimal surface $dS$ which normal $n$ points towards the fluid that is acting on it, the resulting force $df$ can be written as:
\begin{align}
    df = \sigma(n)dS
\end{align}

Let us take an infinitesimal cube, we look only at three contiguous faces, and call $\sigma_{ij}$ the force per unit of surface acting on the $i^{th}$ face on the $j^{th}$ direction. We then define the second order \emph{stress tensor} (also known as the \emph{Cauchy stress tensor}) $\sigma$ whose components are $\sigma_{ij}$. By the third law of Newton, we find that, at first order, for an infinitesimal cube, $\sigma(n) = -\sigma(-n)$ which tells us that we only need to know the tensor $\sigma$ to know completely the surface forces acting on the entire cube (and not only on the three contiguous faces chosen previously). Moreover, by an argument on the kinetic moment of this infinitesimal cube, we find that $\sigma$ needs to be symmetric. 

We conclude that for an arbitrary normal $n$, we only need to project the stress tensor on this normal to have the force acting on it, we find:
\begin{align}
    \sigma(n) = \sigma \cdot n
\end{align}

In the case of a fluid with a null velocity field, there are only normal stresses acting on our infinitesimal cube. Moreover, those stresses are isotropic. We call \emph{pressure}, denoted by $p$, the intensity of those stresses. We then find $\sigma = -pI$, where $I$ is the identity matrix of dimension 3. The minus sign is because we took the convention of outward normal when we defined $\sigma(n)$.

In the case of a general velocity field, we define the \emph{viscous stress tensor} $\sigma'$ by:
\begin{align}
    \sigma = -pI + \sigma'
\end{align}
We remark that $\sigma'$ is also a second order symmetric tensor.

On the other hand, the deformation of an infinitesimal volume of control can be quantified thanks to the Jacobian $J$ of the velocity. This Jacobian can be decoupled in a symmetric tensor $S$ and a skew-symmetric tensor $W$ via:
\begin{align}
    J &= S + W \\
    S &= \frac{1}{2}(J + J^t) \\
    W &= \frac{1}{2}(J - J^t)
\end{align}
where $J^t$ correspond to the transpose of $J$. 

The tensor $W$ represents the pure rotation of the volume of control and the tensor $S$ represents the compression and dilation of the volume of control with respect to a certain basis (as it is symmetric, it can be diagonalizable in an orthonormal basis). Moreover, we remark that $\nabla\cdot v =  \Tr(J) = \Tr(S)$, hence, if the fluid is incompressible, $J$, $S$ and $W$ are traceless. 

For newtonian and incompressible fluids, we find the relation:
\begin{align}
    \sigma' = 2\mu S
\end{align}
where $\mu$ is the coefficient that quantifies the dissipation property of the fluids through shear stresses, we call it the \emph{dynamic viscosity}.

Hence, we find that the stress force $df$ acting on a face of area $dS$ and normal $n$ of our infinitesimal cube is:
\begin{align}
    df = -pn + 2\mu S\cdot n
\end{align}
And we can conclude that for a geometry of surface $\mathcal{S}$, the stress force $F$ acting on it can be computed via:
\begin{align}
    F &= \oint_\mathcal{S} \sigma\cdot n dS \\
    &= -\oint_\mathcal{S} pndS + \oint_\mathcal{S} 2\mu S\cdot n dS
\end{align}
We call the term $P := -pn$ the wall pressure and the term $\tau := 2\mu S\cdot n$ the wall shear stress. Ultimately, we call $drag$ $D$ and $lift$ $L$ the component of $F$ that are respectively parallel and orthogonal to the main direction of the flow. If the fluid flows in the $x$-direction, we have:
\begin{align}
    D &= \left(\oint_\mathcal{S} PdS + \oint_\mathcal{S} \tau dS\right)_x \\
    L &= \left(\oint_\mathcal{S} PdS + \oint_\mathcal{S} \tau dS\right)_y
\end{align}

In the case of RANS equations, we add terms that take in account the effect of turbulence over the geometry. The pressure $p$ is replaced by an effective pressure $p_e$ and the wall shear stress is given by $\tau = 2(\mu + \mu_t)S\cdot n$ where $\mu_t$ is the dynamic \emph{turbulent} viscosity. 

For incompressible fluids we also often divide those quantities by $\rho$ the density of the fluid and solvers often express the results in terms of reduced pressure $p' := p_e/\rho$ and kinematic (turbulent) viscosity $\nu := \mu/\rho$ ($\nu_t := \mu_t/\rho$). We use this convention in this work.

\section{Non-dimensionalization of the Navier-Stokes equation}\label{ap:NS_dim}
Solving Navier-Stokes' equations for a set of parameters $\rho$ and $\nu$ and boundary conditions may actually be equivalent to solving a whole family of equations. To enlighten this phenomena we can work with non-dimensional quantities. Moreover, such formulation will help us to see the importance of each term in the system of partial differential equations. 

In order to do so, let $T$, $L$, $V$, $P$ be characteristics time scale, length scale, velocity scale and pressure scale (respectively) of the problem. We define:
\begin{align}
    t = T\hat{t} \qquad r = L\hat{r} \qquad v = V\hat{v} \qquad p = P\hat{p}
\end{align}
All the quantities with a hat are dimensionless and we can update the incompressible Navier-Stokes equations without source term:
\begin{align}
    \frac{V}{T}\partial_{\hat{t}} \hat{v} + \frac{V^2}{L}(\hat{v}\cdot\hat{\nabla})\hat{v} = -\frac{P}{\rho L}\hat{\nabla}\hat{p} + \frac{\nu V}{L^2}\hat{\Delta}\hat{v}
\end{align}
Let us take $T$ equal to $L/V$ and $P$ equal to $\rho V^2$, we find:
\begin{align}
    \frac{V^2}{L}\partial_{\hat{t}} \hat{v} + \frac{V^2}{L}(\hat{v}\cdot\hat{\nabla})\hat{v} = -\frac{V^2}{L}\hat{\nabla}\hat{p} + \frac{\nu V}{L^2}\hat{\Delta}\hat{v}
\end{align}
In total, this gives:
\begin{align}
    \partial_{\hat{t}} \hat{v} + (\hat{v}\cdot\hat{\nabla})\hat{v} &= -\hat{\nabla}\hat{p} + \frac{1}{Re}\hat{\Delta}\hat{v} \\
    Re &= \frac{VL}{\nu}
\end{align}
The dimensionless number $Re$ is called the \emph{Reynolds number}. This equation only depends on the Reynolds number and two flows with the same Reynolds number will have the same dimensionless solution. 

Moreover, the Reynolds number can be seen as the ratio of the order of magnitude of the convective term over the order of magnitude of the viscous term:
\begin{align}
    Re = \frac{\text{convective term}}{\text{viscous term}} = \frac{\|(v\cdot\nabla)v\|}{\|\nu\Delta v\|} = \frac{V^2/L}{\nu V/L^2} = \frac{VL}{\nu}
\end{align}
We then have two particular regimes:
\begin{itemize}
    \item $Re \to 0$, viscous term dominates the flow (we call this a Stokes flow)
    \item $Re \to \infty$, convective term dominates the flow (Navier-Stokes equations tends towards Euler's equations for inviscid fluids)
\end{itemize}

\end{document}